\documentclass[final,5p,times,twocolumn,]{elsarticle}

\usepackage{amssymb}
\usepackage[table, dvipsnames]{xcolor}
\usepackage{soul}
\usepackage{amsmath}
\usepackage{algorithm}
\usepackage[noend]{algpseudocode}
\usepackage{tabularx}
\usepackage{natbib}
\usepackage{caption}
\usepackage{subcaption}
\usepackage{float}
\usepackage{multirow}
\usepackage{dblfloatfix}
\usepackage{hyperref}
\usepackage{pifont}
\usepackage{pgfplots}
\pgfplotsset{compat=1.17}
\usepackage{array}
\usepackage{makecell}
\usepackage[utf8]{inputenc}
\usepackage[T1]{fontenc}

\usepackage{arydshln} 
\usepackage{booktabs}   
\usepackage{multirow}   
\usepackage{graphicx}   
\usepackage{xcolor}     
\usepackage{pifont}     
\usepackage{amsmath}    
\usepackage{array}      
\usepackage[left]{lineno}  

\journal{Elsevier}

\begin{document}
\begin{frontmatter}

\title{LiDAR Remote Sensing Meets Weak Supervision: Concepts, Methods, and Perspectives}

\author[label1,label2,label3]{Yuan~Gao\corref{eq1}}
\author[label4]{Shaobo~Xia\corref{eq1}}
\author[label5]{Pu~Wang}
\author[label1,label2,label3]{Xiaohuan~Xi}
\author[label1,label2,label3]{Sheng~Nie\corref{cor1}}
\author[label1,label2,label3]{Cheng~Wang\corref{cor1}}
\ead{wangcheng@aircas.ac.cn}

\cortext[eq1]{These authors contributed equally to this work.}
\cortext[cor1]{Corresponding author.}

\affiliation[label1]{organization={Key Laboratory of Digital Earth Science},
            addressline={Aerospace Information Research Institute, Chinese Academy of Sciences}, 
            city={Beijing 100094},
            country={China}}
\affiliation[label2]{organization={International Research Center of Big Data for Sustainable Development Goals},
            city={Beijing 100094},
            country={China}}
\affiliation[label3]{organization={University of Chinese Academy of Sciences},
            city={Beijing 100094},
            country={China}}
\affiliation[label4]{organization={The Department
of Geomatics Engineering, Changsha University of Science and Technology},
            city={Hunan 410004},
            country={China}}

\affiliation[label5]{organization={Institute of Energy, Environment, and Economy, Tsinghua University},
            city={Beijing  100084},
            country={China}
            }            

\begin{abstract}

\textcolor{black}{Light detection and ranging (LiDAR) remote sensing encompasses two major directions: data interpretation and parameter inversion. However, both directions rely heavily on costly and labor-intensive labeled data and field measurements, which constrains their scalability and spatiotemporal adaptability. Weakly Supervised Learning (WSL) provides a unified framework to address these limitations. This paper departs from the traditional view that treats interpretation and inversion as separate tasks and offers a systematic review of recent advances in LiDAR remote sensing from a unified WSL perspective. We cover typical WSL settings—including incomplete supervision (e.g., sparse point labels), inexact supervision (e.g., scene-level tags), inaccurate supervision (e.g., noisy labels), and cross-domain supervision (e.g., domain adaptation/generalization)—and corresponding techniques such as pseudo-labeling, consistency regularization, self-training, and label refinement, which collectively enable robust learning from limited and weak annotations. We further analyze LiDAR-specific challenges (e.g., irregular geometry, data sparsity, domain heterogeneity) that require tailored weak supervision, and examine how sparse LiDAR observations can guide joint learning with other remote-sensing data for continuous surface-parameter retrieval. Finally, we highlight future directions where WSL acts as a bridge between LiDAR and foundation models to leverage large-scale multimodal datasets and reduce labeling costs, while also enabling broader WSL-driven advances in generalization, open-world adaptation, and scalable LiDAR remote sensing.}

\end{abstract}


\end{frontmatter}

\section{Introduction}

\textcolor{black}{Light detection and ranging (LiDAR)}, an active remote-sensing modality typically based on laser time-of-flight ranging, has emerged as a cornerstone of modern remote sensing, offering exceptional capabilities in capturing high-precision 3D data of the Earth’s surface \cite{CHEN2024181,Wang2024IntroductionTL}. \textcolor{black}{LiDAR remote sensing has reshaped geospatial analysis, supporting applications in terrain mapping, ecological monitoring, agricultural and forestry resource management, cryosphere observations, urban modeling, as well as infrastructure inspection and hazard assessment.}

\begin{figure*}[t!]
    \centering
    \includegraphics[scale = 0.1]{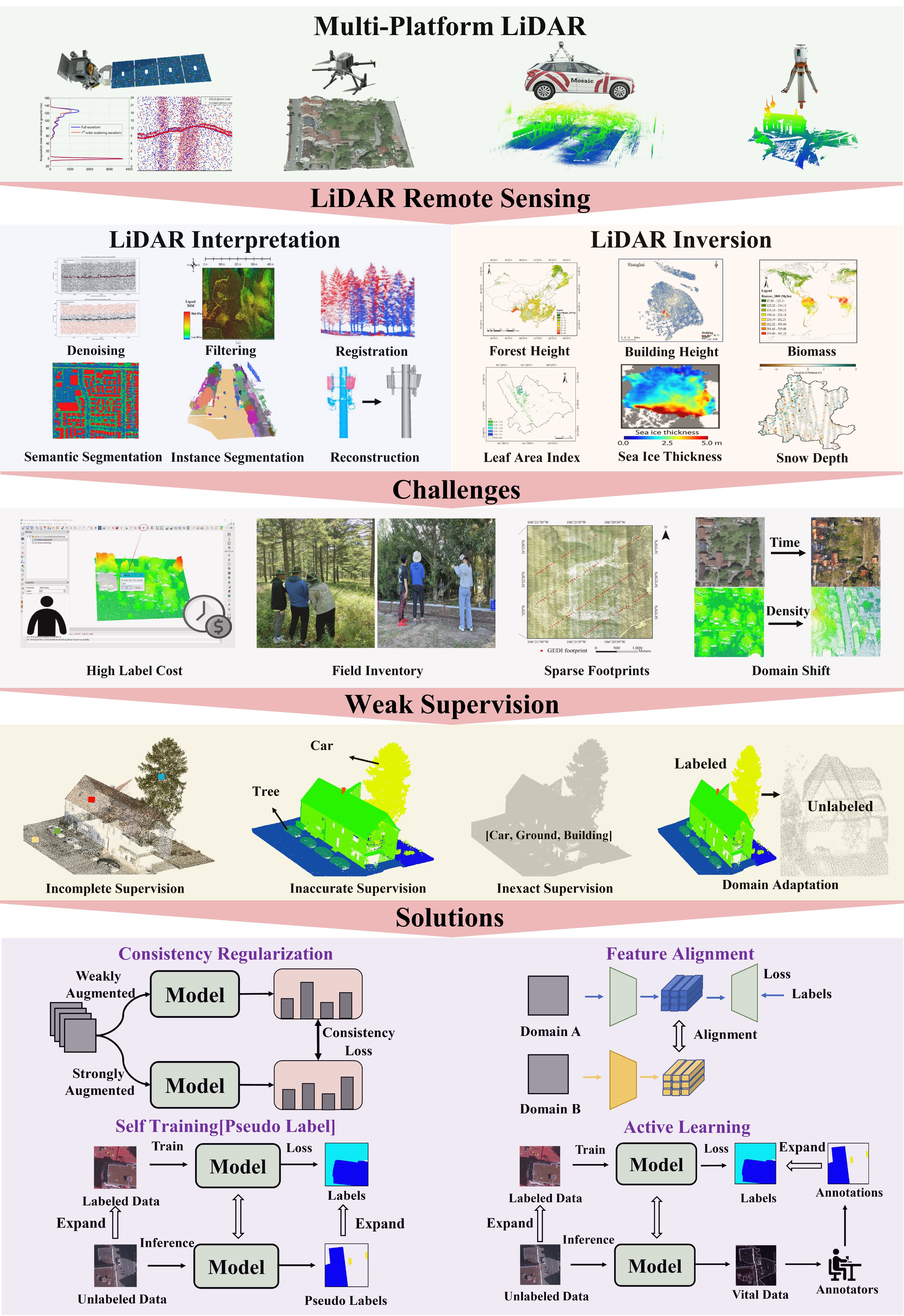}
    \caption{\textcolor{black}{An overview of LiDAR remote sensing meets weak supervision. LiDAR remote sensing data are primarily obtained from satellite, airborne, vehicle-mounted, and ground-based platforms, providing rich surface information. The interpretation tasks for LiDAR data include denoising, filtering, registration, semantic segmentation, instance segmentation, and 3D reconstruction. Inversion tasks involve estimating forest height, building height, biomass, leaf area index, sea ice thickness, and snow depth. These tasks face challenges such as high annotation costs, domain shift, difficulties in field validation, and sparse supervisory signals. To address these challenges, weakly supervised learning techniques such as incomplete, inaccurate, inexact supervision, domain generalization, and domain adaptation can be employed. Additionally, methods like consistency regularization, pseudo-labeling, self-training, active learning, and feature alignment can effectively improve model generalization. These images are
obtained from \cite{Wang2024IntroductionTL,duan2023denoising,KOLLE2021100001,HAN2024500,f15071257,BESSO2024113843,rs16162983,9150622,10636266,10679232,qiao2024framework,GUENTHER2024114293}.}}
    \label{fig:overframe}
\end{figure*}

From an interdisciplinary perspective, LiDAR remote sensing plays a dual role across diverse applications: it serves as a data source requiring interpretation while also acting as a sparse signal supporting large-scale remote sensing inversion. \textcolor{black}{As a data source, LiDAR measurements are collected from multiple platforms, typically including spaceborne systems (e.g., Global Ecosystem Dynamics Investigation (GEDI) and Ice, Cloud, and land Elevation Satellite-2 (ICESat-2)), airborne laser scanning (ALS), terrestrial laser scanning (TLS), and mobile laser scanning (MLS). These platforms collect massive volumes of high-resolution 3D data, which cannot be directly applied in downstream analyses. Instead, the raw data must undergo interpretation—typically involving preprocessing (e.g., denoising~ \cite{zhu2020noise,duan2023denoising}), filtering~ \cite{10636266}, classification \cite{9915611}, segmentation~ \cite{9028090}, and reconstruction~ \cite{qiao2024framework}—to derive meaningful geometric and semantic information that supports analytical and operational applications.}
\textcolor{black}{As a sparse signal, LiDAR provides physically grounded, high-quality observations that serve as critical constraints or weak supervision for large-scale land surface parameter inversion. Termed LiDAR-based inversion, this involves exploiting LiDAR data and machine learning to map sparse yet physically grounded observations to geophysical and geometric parameters (e.g., building height \cite{MA2023113392}, canopy height \cite{pmlr-v235-pauls24a}, leaf area index~\cite{f15071257}, biomass \cite{rs14143432}, sea ice thickness~\cite{rs16162983}, and snow depth~\cite{BESSO2024113843}), thereby enabling large-area parameter retrieval from limited reference measurements, as illustrated in Figure~\ref{fig:overframe}.}

LiDAR data interpretation and LiDAR-based inversion face several common challenges. Firstly, LiDAR data interpretation often relies on a large number of accurate manual annotations, but the high quality of such annotations requires significant time and resources, making the process costly in practical applications \cite{xia2023densify}. Similarly, LiDAR-based inversion also encounters high annotation costs. Due to the sparsity of LiDAR data, large-scale, continuous surface mapping is difficult to achieve independently \cite{lang2023high}. However, acquiring more annotated data typically necessitates field surveys to measure diameter at breast height, tree height, leaf area index, etc., which is time- and resource-intensive \cite{TOLAN2024113888}. Secondly, both LiDAR data interpretation and LiDAR-based inversion face domain shift issues \cite{Wang2024TesttimeAF,brandt2024high}. LiDAR data collected from different platforms exhibit significant variations in density, resolution, coverage, and noise levels, leading to notable domain shifts. Additionally, the Earth's surface is diverse and constantly changing, covering various terrains such as cities, forests, mountains, and water bodies, which further exacerbates domain shift problems.

\textcolor{black}{In response to these challenges, weakly supervised learning (WSL) provides an effective solution \cite{WANG2022237}. WSL is a broad machine learning paradigm that constructs predictive models under limited, weak, or mismatched supervision, encompassing settings with incomplete, inexact, or noisy labels as well as cross-domain scenarios such as domain adaptation and domain generalization. Unlike fully supervised learning, which relies on large amounts of precise annotations, weakly supervised approaches can leverage sparse, uncertain, or heterogeneous supervision (e.g., crowd-sourced annotations) to substantially reduce annotation costs while maintaining generalization. Building on this extended taxonomy, We adapt WSL as a machine learning paradigm for LiDAR remote sensing, in which models leverage sparse, coarse, noisy, or domain-shifted supervisory signals—together with intrinsic data structures and explicit priors—to learn representations and inference mechanisms that support dense and accurate data interpretation and inversion.
In the context of LiDAR data interpretation, weak supervision enables the extraction of meaningful geometric and semantic information (e.g., structural modeling and semantic segmentation) from limited labels through techniques such as consistency regularization \cite{xu_weakly_2020}, self-training \cite{li_semi-supervised_2021}, and pseudo-labeling \cite{sohn2020simple}. For LiDAR-based inversion, we treat LiDAR measurements as sparse yet physically grounded supervision that anchors large-scale parameter retrieval: models fuse dense imagery with these incomplete LiDAR observations to learn a sparse-to-dense mapping from observations to geophysical and biophysical parameters.}

However, most researchers have focused solely on LiDAR data interpretation under weak supervision or only on LiDAR-assisted large-scale remote sensing inversion. Few studies have been able to examine LiDAR remote sensing from a unified perspective, considering the dual role of LiDAR.
Although several reviews on point clouds have been published, they have certain limitations.
\textcolor{black}{The survey \cite{10561540} systematically examines key methodologies such as data augmentation, domain adaptation, WSL, and pre-trained foundation models, discussing research progress in tasks like object detection, classification, and segmentation across synthetic datasets, indoor scenes, and autonomous driving scenes. Notably, the scope of this review excludes remote sensing data (e.g., ALS and spaceborne photon-counting point clouds) and remote sensing inversion tasks, such as the estimation of canopy and building heights. 
\textcolor{black}{Guo et al. \cite{9127813} and He et al. \cite{HE2025102722} provide a comprehensive review for 3D point-cloud processing, covering classification, detection, and segmentation. However, their discussion of techniques is largely confined to fully supervised methods and overlooks the potential of WSL. Moreover, their review omits the increasingly important area of remote-sensing LiDAR point clouds.}
Xiao et al. \cite{10086697} systematically reviews various unsupervised deep learning methods that learn general and robust feature representations by performing pretext tasks, such as generative or contrastive learning, on large-scale unlabeled 3D point clouds. These learned representations can then be effectively transferred to downstream supervised tasks. However, the survey does not sufficiently discuss the role and potential of WSL in LiDAR remote sensing. Therefore, a systematic and comprehensive review of LiDAR remote sensing from a weakly supervised perspective is both necessary and timely.}

\begin{figure*}[tbhp]
    \centering
    \includegraphics[scale = 0.18]{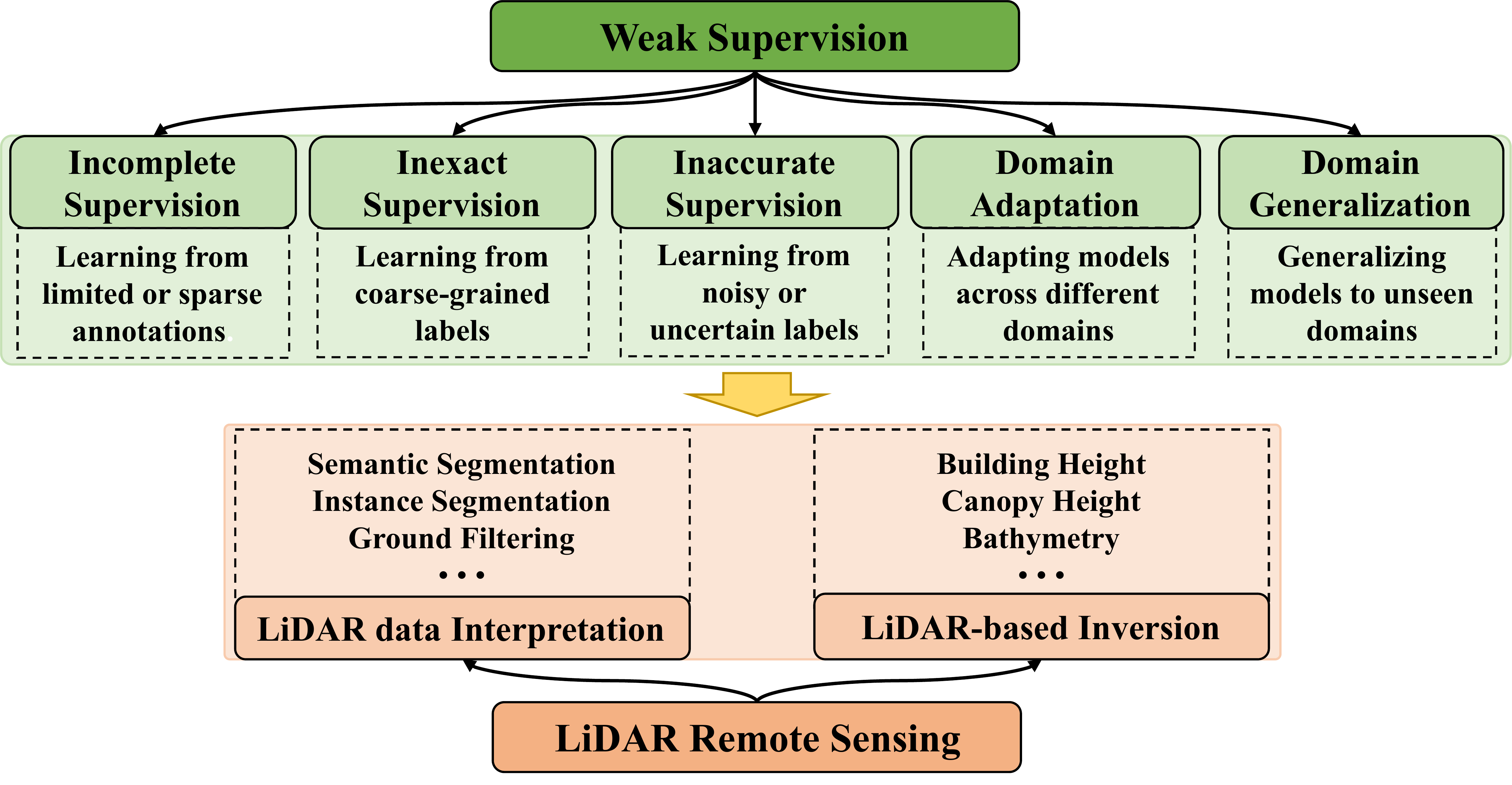}
    \caption{\textcolor{black}{A taxonomy of methods and applications for LiDAR remote sensing meets weak supervision. Weakly supervised approaches for LiDAR remote sensing include incomplete supervision, inexact supervision, inaccurate supervision, domain adaptation and domain generalization. These approaches jointly support point cloud interpretation and LiDAR-based inversion tasks. 
}}
    \label{fig:framework}
\end{figure*}

To bridge the gaps between different research directions and advance the development of LiDAR remote sensing, this review focuses on two primary aspects: \textbf{the interpretation of LiDAR remote sensing data under weak supervision} and \textbf{the use of LiDAR as a weak supervisory signal in remote sensing}. It aims to comprehensively explore the potential and challenges of LiDAR remote sensing in various applications from the perspective of weak supervision. 
Fig. \ref{fig:framework} provides an overview of the entire article. The contributions of this paper are summarized as follows:
\begin{itemize}
\setlength{\parskip}{3pt}
    \item To the best of our knowledge, this is the first review to examine LiDAR remote sensing from a unified perspective, considering its dual role as both a data source and a weak supervisory signal.\par
    
    \item This paper explores recent advancements in interpreting LiDAR data under weak supervision and utilizing LiDAR as a weak supervisory signal in remote sensing, offering readers state-of-the-art methods.

    \item This paper highlights promising research directions in LiDAR remote sensing from a weak supervision perspective and explores the potential for integrating more weakly supervised learning techniques into LiDAR-based inversion.
\end{itemize}

\textcolor{black}{This paper is organized as follows. Section \ref{section_preliminaries} provides the background, outlining specialized deep learning architectures for LiDAR point clouds and the rationale for the paradigm shift from fully to weakly supervised learning. Sections \ref{section_interpretation} and \ref{section_signals} form the core of the paper, reviewing the applications of weakly supervised learning in LiDAR data interpretation and parameter inversion, respectively. Section \ref{section_future} offers a forward-looking perspective, analyzing the unique characteristics of LiDAR data, the limitations of current methods, the potential of foundation models, and future research directions. Finally, Section \ref{section_conclusion} concludes the paper.}

\section{Preliminaries}\label{section_preliminaries}

\subsection{Deep learning for LiDAR}

\textcolor{black}{LiDAR data primarily include point clouds acquired from ground-based, airborne, and spaceborne platforms, as well as large-footprint full-waveform measurements. To process the massive amount of LiDAR data, deep learning–based methods are commonly employed to extract semantic and geometric information for downstream applications. However, point cloud data are inherently unstructured, unordered, and distributed in three-dimensional space, which makes conventional image-based neural network architectures inapplicable. This stems from the fact that these architectures—such as convolutional and transformer-based networks—rely on regular grid structures to enable local feature aggregation and position encoding. In contrast, point clouds lack such fixed adjacency and ordering, making it difficult to directly apply conventional deep learning architectures designed for regular grids. }

\textcolor{black}{Therefore, to accommodate the unstructured, unordered, and highly sparse nature of point clouds, researchers have developed network families tailored to 3D geometry rather than regular image grids. Point-based models (e.g., PointNet \cite{qi2017pointnet}/PointNet++ \cite{qi2017pointnet++}) operate directly on 3D coordinates with permutation-invariant set functions and local neighborhood aggregation; graph-based models (e.g., DGCNN \cite{wang2019dynamic}) build dynamic kNN/radius graphs to capture adaptive local topology; kernel/stateful convolutions and efficient sampling (e.g., KPConv \cite{thomas2019kpconv}, RandLA-Net \cite{hu2020randla}) improve locality modeling and scalability on large scenes. In parallel, voxel-based pipelines (e.g., VoxelNet \cite{zhou2018voxelnet}, submanifold/Minkowski sparse 3D CNNs \cite{choy20194d}) discretize space and leverage sparse tensors for efficient convolutions. More recently, Transformer-style architectures (e.g., Point Transformer \cite{zhao2021point} and its variants) employ self-attention to model long-range dependencies and cross-scale interactions, often combined with local grouping or sparse attention for efficiency. Most of these architectures operate directly on irregular 3D points, explicitly handling nonuniform density, occlusions, and sparsity to achieve efficient and robust 3D representation learning. For a detailed review for 3D deep learning, we refer the reader to \cite{guo2020deep} and \cite{he2025deep}. }

\subsection{From fully to weakly supervised learning}
\textcolor{black}{With the rapid development of neural architectures for LiDAR data processing, training robust, accurate, and generalizable deep learning models for point cloud data has become a central challenge. Most existing approaches rely on fully supervised learning, a paradigm in which models are trained on densely labeled datasets—each 3D point is assigned a ground-truth class or attribute, and the learning objective minimizes the discrepancy between predictions and reference labels. This setting enables direct end-to-end optimization and often achieves high predictive accuracy. Fully supervised learning has led to significant advances in 3D scene understanding. For example, RandLA-Net achieved 83.1\% overall accuracy (OA) and 70.1\% mF1 on ISPRS; KPConv reported over 81\% mIoU on DALES.}

\textcolor{black}{However, unlike 2D image tasks—where annotations can often be obtained at low cost from web data or automatically generated using vision-language models—fully supervised training for LiDAR data faces prohibitively high labeling costs. For point cloud interpretation, annotations rely heavily on manual labeling with specialized 3D software, which is labor-intensive, error-prone, and difficult to scale. For LiDAR-based inversion, collecting ground reference data is expensive, requires intensive fieldwork, and may pose safety risks. Moreover, LiDAR remote sensing targets the Earth’s surface, which is vast, heterogeneous, and continuously changing due to seasonal dynamics, human activities, and natural disturbances. These factors make it extremely difficult to obtain high-quality, spatiotemporally comprehensive annotations, rendering fully supervised training unsustainable at scale.}

\textcolor{black}{To tackle these challenges, weakly supervised learning (WSL) has emerged as a promising paradigm in LiDAR remote sensing. Instead of relying on exhaustive point-level annotations, WSL leverages limited or imprecise labels to train deep learning architectures specialized for point clouds, thereby substantially reducing the dependence on large-scale manual labeling in data interpretation. In the context of LiDAR-based inversion, WSL enables the use of sparse field measurements collected over limited areas to support large-scale parameter retrieval. By incorporating incomplete supervision and promoting domain generalization, weak supervision can mitigate the effects of spatial heterogeneity and temporal variability, allowing models trained on sparse local data to generalize effectively across broader regions. Consequently, WSL provides a practical and scalable alternative to conventional supervised paradigms and represents a critical research direction for advancing both LiDAR data interpretation and parameter inversion.}

\begin{table*}[htbp] %
	\scriptsize
	\centering %
	\caption{\textcolor{black}{A summary of commonly used semantic segmentation datasets for LiDAR remote sensing}}
	\resizebox{0.8\textwidth}{!}{%
		\centering
		\begin{tabular}{lccccc} %
			\toprule %
			
			Dataset & Platform & Year & Area/Length & \# Classes & \# Points  \\ %
			\midrule %
			
			Semantic3D~\cite{hackel2017isprs} & \multirow{1}{*}{ Terrestrial Laser Scanning} & 2017 & - & 8 & 4000M   \\
			\midrule
			
			Okland~\cite{5206590} & \multirow{7}{*}{Mobile Laser Scanning} & 2009 & $1.5 \times 10^3 \: m$ & 5 & 1.6M  \\
			Paris-rue-Madame~\cite{serna2014paris} &  & 2014 & $0.16 \times 10^3 \: m$  & 26 & 20M  \\
			iQmulus~\cite{vallet2015terramobilita} &  & 2015 & $ 0.2 \times 10^3 \: m$ & 22 & 12M  \\
			TUM-City-campus~\cite{rs12111875} &  & 2016 & $0.97 \times 10^3 \: m$ & 8 & 41M  \\
			Paris-Lille-3D~\cite{8575440} &  & 2018 & $1.94 \times 10^3 \: m$ & 9 & 143M  \\
			Toronto-3D~\cite{9150609} &  & 2020 & $1 \times 10^3 \: m$ & 8 & 78.3M   \\
			WHU-Urban3D~\cite{HAN2024500} &  & 2023 & $6.5 \times 10^3 \: m$ & 18(35) & 393M   \\
			\midrule
			
			ISPRS~\cite{isprs-dataset} & \multirow{8}{*}{Aerial Laser Scanning} & 2012 & - & 9 & 1.2M   \\
			DublinCity~\cite{zol2019dublincity} &  & 2019 & $2 \times 10^6 \: m^2$ & 13 & 260M   \\
			DALES~\cite{9150622} &  & 2020 & $10 \times 10^6 \: m^2$ & 8 & 505M  \\
			LASDU~\cite{ijgi9070450} &  & 2020 & $1.02 \times 10^6 \: m^2$ & 5 & 3.12M  \\
			Hessigheim 3D~\cite{KOLLE2021100001} &  & 2021 & $0.19  \times 10^6 \: m^2$ & 11 & 125.7M/36.76M  \\
			WHU-Urban3D~\cite{HAN2024500} &  & 2023 & $3.2 \times 10^6 \: m^2$ & 9 & 213M   \\
			ECLAIR~\cite{10678007} &  & 2024 & $10.3 \times 10^6 \: m^2$ & 11 & 582M  \\
			FRACTAL~\cite{gaydon2024fractal} &  & 2024 & $250  \times 10^6 \: m^2$ & 7 & 9261M  \\
			\bottomrule %
		\end{tabular}
	}
	\label{tbl:dataset_comparison}
\end{table*}

\begin{table*}[htbp]
	\centering
	\renewcommand\arraystretch{1.0}
	\caption{\textcolor{black}{Representative works of LiDAR remote sensing interpretation with weak supervision.}}
	\vspace{-3mm}
	\label{tab:overview}
	\footnotesize
	\setlength{\tabcolsep}{3.5pt}{
		\resizebox{\textwidth}{!}{
			\begin{tabular}{c|c|c|c} 
				\hline
				\textbf{Weak supervision} & \textbf{Setting} & \textbf{Method} & \textbf{Contribution} \\ \hline
	\multirow{33}{*}{\makecell{Incomplete \\ Supervision}} & \multirow{6}{*}{\makecell{Semi \\ supervision}}  
				& SSGF\cite{7325695} & Handle labeled and unlabeled data via graphs separately \\  
				&    & Li et al. \cite{10660507} & Enhance intra-class information interaction via contrastive learning\\  
				&         & TSDN \cite{9915611} &  Generate high-confidence pseudo-labels using superpixel segmentation and multiview prediction results\\  
				&       & UF2SCN \cite{Guo2023SemisupervisedCF} & Improve the complementarity of HSI and LiDAR data via a one-way transmission structure \\  
				&       & SMDN \cite{10609507} & Guide feature training in a subspace with limited labeled samples \\  
				&           & Dersch et al. \cite{Dersch2024SemisupervisedMT} & Generate tree labels by pre-clustering LiDAR point clouds using NCut \\ 
				
				\cline{2-4}   
		& \multirow{12}{*}{\makecell{Sparse \\ supervision}}          
				&   Wang et al. \cite{isprs-annals-V-2-2021-43-2021}    &   Select reliable pseudo-label via adaptive thresholding  \\   
				&    & Wang et al. \cite{WANG2022237} & Penalize class overlap in predictive probabilities via entropy regularization \\ 
				& &PSD \cite{zhang_perturbed_2021} & Establish point cloud graph topology to enable information propagation between labeled and unlabeled points\\
				& &Zhang et al. \cite{Zhang2021WeaklySS} & Transfer prior knowledge from unlabeled data to weakly supervised networks via self-supervised point cloud colorization pre-training \\
				& &SQN \cite{hu2022sqn} & Enhance sparse (0.1\%) annotations using semantic homogeneity in point cloud neighborhoods \\
				&    & WSPointNet \cite{lei_wspointnet_2022} & Provide adaptive supervision through a pseudo-label branch \\ 
				& & MSCs\cite{10292660} & Apply scene-level constraints at the encoder, decoder, and classifier stages to alleviate class imbalance \\ 
				& &Chen et al. \cite{10696960}
				& Use line and plane points as additional supervisory signals for hierarchical feature learning\\
				& &FPWS-Net \cite{chen_feature_2024}
				& Enforce consistency on unlabeled data via mutual pseudo-labeling loss \\
				& & SCSQ-Net \cite{10767226} & Share weights between encoder and semantic inference to efficiently integrate neighborhood context \\
				& & Li et al. \cite{LI2024110701} & Decouples feature learning and the classifier via alternating optimization \\
				& & HSCN \cite{10735245} & Enforce global distribution alignment with GAL and multiscale neighborhood-structure alignment with similarity metric loss \\
				& & DR-Net \cite{10488461}  & Enhance feature representation by aggregating local features of the point cloud \\
				& & DSDCL \cite{10855480} & Employ context-awareness to enhance the model's ability to differentiate similar points \\
				& & DCUF-Net \cite{10813603} & Use an uncertainty-aware loss to extract reliable supervisory signals from large unlabeled datasets \\  \cline{2-4}  
	& \multirow{10}{*}{\makecell{Active learning}}          
				& Polewski et al. \cite{7378854}  &  Combine semi-supervised and active learning under Rényi entropy regularization \\ 
				& &  Luo et al. \cite{8353504}   & Employ an active learning strategy to iteratively annotate a limited number of supervoxels  \\
			& 	&  Lin et al. \cite{isprs-annals-V-2-2020-243-2020}   & Propose a segmentation-based active learning strategy that iteratively annotates informative unlabeled samples  \\
			& 	&   Lin et al. \cite{LIN202073}   &  Use point entropy, segmentation entropy, and mutual information to iteratively select the most informative samples \\
			& 	&  Kölle et al. \cite{isprs-annals-V-2-2021-93-2021}   & Combine active learning with crowdsourced annotation  \\
			& 	&  SSDR-AL \cite{10.1145/3503161.3547820}   & Select the most informative and representative samples through a graph reasoning network  \\
			& 	&  Kölle et al. \cite{isprs-annals-X-1-W1-2023-945-2023}    &  Validate Active Learning performance \\
			& 	&   Wang et al. \cite{wang2023one}  & Integrate the novel 'One-Class-One-Click' method with uncertainty-driven active learning  \\
			& 	&  Dai et al. \cite{10788411}   & Introduce an Active Learning strategy with a Spatially Incremental Sampling method for label acquisition \\
			&    &  DAAL-WS \cite{LEI2024103970} & Select critical labeling points via active learning \\ 
			\cline{2-4} 
				
	& \multirow{5}{*}{\makecell{Few-shot learning}}          
				&  Thr-MPRNet \cite{10430180}    &  Construct a 3D few-shot framework with feature and relation learners \\
			& 	&  Huang et al. \cite{10423773}   &  Combine few-shot learning with distribution calibration \\
			& 	&   Zhang et al. \cite{isprs-annals-X-4-W5-2024-333-2024}  &  Introduce Pole-NN, a training-free, non-parametric feature extractor \\
			& 	&  Dai et al. \cite{10818527}   &  Decompose cross-domain ALS segmentation into transfer learning and class-incremental learning \\
			& 	&  Yuan et al. \cite{rs17152618}   & Propose a Multi-branch Feature Extractor with three-stage transfer learning: pre-training, cross-domain alignment, and incremental learning  \\
				\hline                                    
	\multirow{5}{*}{\makecell{Inexact \\ Supervision}}     & Box level supervision     &   - &-    \\ \cline{2-4}
				& Subcloud level supervision&   Lin et al. \cite{LIN202279}                 &   Explore semantic information between adjacent subclouds through overlap region loss          \\ \cline{2-4} 
			& \multirow{2}{*}{\makecell{Scene level supervision}}  &      Wang et al. \cite{wang2023one}       &  Select informative subclouds for annotation using TOD-based active learning   \\
				& & WHCN \cite{LU2025129264}  & Construct a hypergraph from superpoints to generate point-level pseudo-labels\\
				 \cline{2-4}
				& Scribble level supervision &      -          &     -     \\ \hline
		Inaccurate Supervision                   & Noisy label supervision   & ADNLC \cite{10677356}  &  Help the network learn true labels by iteratively updating noisy labels    \\ \hline
	\multirow{12}{*}{\makecell{Domain \\ Adaptation}}   
		   & \multirow{9}{*}{\makecell{Unsupervised \\ domain adaptation}}  
				&   Luo et al. \cite{Luo2020UnsupervisedSA}      &     Align source and target domain feature distributions using an MCD-based adversarial framework     \\  
				&        &   Peng et al. \cite{peng2020point}      &     Achieve point feature and global feature alignment through adversarial training     \\ 
				&          &   Xie et al. \cite{Xie2021EXPLORINGCS}      &   Reduce cross-city domain shift via appropriate point cloud density and suitable data augmentation methods  \\ 
				&          &  SegTrans \cite{10179997}      &  Align feature distributions of source and target domains using adversarial learning     \\  
				&            &   Wang et al. \cite{Wang2024ADA}      &  Reduce domain gaps using alignment, normalization, enhancement, and weighted cross-entropy     \\  
				& & Wang et al. \cite{10641953}  & Use geometric transformation augmentation to enable cross-modal learning from LiDAR to SAR point clouds\\
				& & PS-UDA \cite{10586968} & Employ an L2-paradigm constraint and a Laplace matrix to pre-align cross-scene features\\
				& & Luo et al. \cite{LUO2025537} & Reconstruct MLS scene surfaces and align source and target point clouds in a canonical domain at the point level\\
				 \cline{2-4}
			&  Source-free domain adaptation    &     -     &     -     \\ \cline{2-4}
			& \multirow{2}{*}{\makecell{Test-time adaptation}}            &    Yao et al. \cite{Wang2024TesttimeAF}      & Update BN stats and affine parameters using test batch statistics and a self-supervised module      \\
				& & APCoTTA \cite{gao2025apcottacontinualtesttimeadaptation}  & Update partial layers and interpolating parameters to retain source knowledge \\
				\hline
		\end{tabular}}
		\label{tbl:weak_supervision}
		\vspace{-5mm}
	}
\end{table*}
\begin{table*}[t!]
	\centering
	\renewcommand\arraystretch{1.0}
	\caption{\textcolor{black}{
    Weakly supervised LiDAR remote sensing interpretation results on different datasets.(``1'' denotes only one point; ``1\%'' indicates that 1\% of points are randomly labeled for each class. ``AL'' stands for active learning, and ``FW'' denotes few-shot learning.
		)}}
	\vspace{-3mm}
	\label{tab:overview}
	\footnotesize
	\setlength{\tabcolsep}{3.5pt}{
		\resizebox{\textwidth}{!}{
			\begin{tabular}{c|c|c|ccc|ccc|ccc|ccc|ccc|ccc|ccc} 
				\hline
				\multirow{2}{*}{\textbf{Method}} & \multirow{2}{*}{\textbf{Backbone}}& \multirow{2}{*}{\textbf{Supervison}} & \multicolumn{3}{c|}{\textbf{ISPRS}}  & \multicolumn{3}{c|}{\textbf{LASDU}}  & \multicolumn{3}{c|}{\textbf{H3D}}  & \multicolumn{3}{c|}{\textbf{DALES}} & \multicolumn{3}{c|}{\textbf{Toronto3D}} & \multicolumn{3}{c|}{\textbf{WHU-MLS}} & \multicolumn{3}{c}{\textbf{Semantic3D}}\\ 
				&       &                                   & OA & mIoU  & mF1   & OA & mIoU  & mF1   & OA & mIoU  & mF1  & OA & mIoU  & mF1 & OA & mIoU  & mF1  & OA & mIoU  & mF1  & OA & mIoU  & mF1  \\
				\hline 
				PointNet++ \cite{qi2017pointnet++}  	&   &  100\%  &81.2  &  & 65.6 &  &  &   &  &  &    &95.7 &68.3 &   &84.88  & 41.81 &  & &  & \\ 
				PointNet++(MSG) \cite{qi2017pointnet++}  	&   &  100\%   &   &  &  &  &  &   &  &  &    &  &  &   &92.56  & 59.47 &  & &  & & &  &  \\ 
				DGCNN \cite{wang2019dynamic}  	&   &  100\%   &   &  &  &  &  &   &  &  &    &  &  &   &94.24  & 61.79  &  & &  & & &  &  \\ 
				RandLA-Net \cite{hu2020randla} & & 100\%  & 83.1&  &70.1 &  &  &   &  &  &    &97.1 &80.0 &   &95.63 & 77.72  &  & &  &  &94.8 &77.4 & \\ 
				KPConv \cite{thomas2019kpconv}  	&   &  100\%  & 84.2  &  & 70.5 &  &  &   &  &  &    &97.8 &81.1 &   &95.39  & 69.11 &  & &  &  &92.9  &74.6& \\ 
				ShellNet \cite{9010996} &   &  100\%  &   &  &  &  &  &   &  &  &    &  &  &   &   &  &  & &  &  &93.2  &69.3& \\ 
				PointGCR \cite{9093411}  &   &  100\%  &   &  &  &  &  &   &  &  &    &  &  &   &   &  &  & &  &  &92.1 &69.5& \\ 
				\hline
				\multirow{4}{*}{Wang et al. \cite{isprs-annals-V-2-2021-43-2021}}   	&   \multirow{4}{*}{KPConv  }   
				& 15  & 80.4  & & 63.1 & & &  & & &  & & &  & & &  & & &  & & &\\ 
				&     & 30  & 81.1  & & 65.4& & &  & & &  & & &  & & &  & & &  & & &\\ 
				&     &  60 &  83.0 &  & 68.6& & &  & & &  & & &  & & &  & & &  & & &\\ 
				&      & 100  & 83.7  &  & 70.2& & &  & & &  & & &  & & &  & & &  & & &\\ 
				\hdashline
				\multirow{5}{*}{Wang et al. \cite{WANG2022237}}		& \multirow{5}{*}{KPConv}   & 0.01\% &  &  &  &  & &  &85.8 & &72.1 & & & & & & & & & & & &\\ 
				&  & 0.05\% & 78.2 &  & 61.9 &87.3 & & 74.6 &86.7 & &75.0& & & & & & & & & & & &\\ 
				&  & 0.1\% & 83.0 &  &70.0 &87.6 & & 77.4 &88.2 & &76.3& & & & & & & & & & & &\\ 
				&   & 0.2\% & 83.4 &  &70.7 &87.7 & & 78.1 &87.5 & & 76.6 & & & & & & & & & & & &\\ 
				&    & 0.5\% & 84.0 &  &71.9 &87.6 &  & 78.3  &  & & & & & & & & & & & & & &\\  \hdashline
				PSD \cite{zhang_perturbed_2021} & RandLA-Net & 1\%  &   &  &  &  &  &   &  &  &    &  &  &   &96.2 &76.2 &   & &  &  & 94.3 &75.8&  \\ \hdashline
				\multirow{2}{*}{Zhang et al. \cite{Zhang2021WeaklySS}} & \multirow{2}{*}{RandLA-Net}  & 1\%  &   &  &  &  &  &   &  &  &    &  &  &   &  &  &   & &  &  & 93.7 &72.6&  \\
				&  & 10\%  &   &  &  &  &  &   &  &  &    &  &  &   &  &  &   & &  &  & 94.0 &73.3&  \\ \hdashline
				\multirow{3}{*}{SQN \cite{hu2022sqn}} &  \multirow{3}{*}{RandLA-Net} & 0.01\%  &   &  &  &  &  &   &  &  &    &95.9 &60.4 &   &94.2 &68.2 &   & &  &   & 90.3 &65.6& \\
				&  & 0.1\%  &   &  &  &  &  &   &  &  &    &97.0 &72.0 &   &96.7 &77.7 &   & &   &  & 93.7 &74.7&  \\ 
				&  & 1\%  &   &  &  &  &  &   &  &  &    &  &  &   &96.9 &80.2 &   & &   &  &  & &  \\ \hdashline
				WSPointNet \cite{lei_wspointnet_2022}	  & RandLA-Net & 0.1\%  &   &  &  &  &  &   &  &  &    &  &  &   &96.76 &78.96 &   & 90.83&56.48  &  &  & &  \\ \hdashline
				MSCs\cite{10292660}	&  RandLA-Net & 1\%  &   &  &  &  &  &   &  &  &    &  &  &   &96.8 &81.2 &   & &   &  &  & &  \\ \hdashline
				\multirow{3}{*}{Chen et al. \cite{10696960}}	& \multirow{3}{*}{KPConv} & 0.05\% &73.5  &  & 55.1 &83.2 &  &71.6  &  &  &    &  &  &   &  &  &   & &   &  &  & &  \\ 
				& & 0.1\%&78.2 &  & 65.6 &84.6 &  &74.5  &  &  &    &  &  &   &  &  &   & &   &  &  & &  \\    
				& & 0.5\% &82.3  &  & 70.1 &84.9 &  &75.3  &  &  &    &  &  &   &  &  &   & &   &  &  & &  \\   \hdashline
				\multirow{2}{*}{Li et al. \cite{LI2024110701}} & \multirow{2}{*}{RandLA-Net} &1pt &   &  &  &  &  &   &  &  &    &  &  &   &  &  &   & &  &  & 92.8 &66.2&  \\ 
				& & 1\% &   &  &  &  &  &   &  &  &    &  &  &   &  &  &   & &  &  & 94.9 &76.9&  \\ \hdashline
				\multirow{2}{*}{HSCN \cite{10735245}}	& \multirow{2}{*}{KPConv} & 0.1\% &79.5  &  & 68.7 &85.2 &  &75.8  &  &  &    &  &  &   &  &  &   & &   &  &  & &  \\ 
				& & 1\% &82.9  &  & 72.8 &86.5 &  &77.7  &  &  &    &  &  &   &  &  &   & &   &  &  & &  \\ \hdashline
				\multirow{2}{*}{DR-Net \cite{10488461}}	& \multirow{2}{*}{SQN \cite{hu2022sqn} } & 0.1\%  &   &  &  &  &  &   &  &  &    &96.6 &72.9 &   &96.9 &78.1 &   & &   &  &  & &  \\ 
				& & 1\% &   &  &  &  &  &   &  &  &   &97.0 &78.0 &   &96.6 &79.1 &   & &   &  &  & &  \\ \hdashline
				\multirow{2}{*}{DSDCL \cite{10855480}}&\multirow{2}{*}{RandLA-Net}& 0.1\% &79.7  &  & 65.7 &85.4 &  &73.4  &  &  &    &  &  &   &  &  &   & &   &  &  & &  \\ 
				& & 1\% &81.8  &  & 70.1 &87.3 &  &77.0  &  &  &    &  &  &   &  &  &   & &   &  &  & &  \\ \hdashline
				DCUF-Net \cite{10813603}	& RandLA-Net& 1\%  &   &  &  &  &  &   &  &  &    &  &  &   &97.2 &81.6 &   & &   &  &  & &  \\  \hline 
				\multirow{2}{*}{Kölle et al. \cite{isprs-annals-X-1-W1-2023-945-2023}} & Random Forest & AL &81.59  &  & 67.46   &  &  &  &84.84 &  &70.39  &  &  &   &  &  &   & &   &  &  & &  \\ 
				& Submanifold Sparse CNN & AL &80.91  &  & 66.20   &  &  &  &85.44 &  &70.90  &  &  &   &  &  &   & &   &  &  & &  \\ \hdashline
				\multirow{2}{*}{Wang et al. \cite{wang2023one}}  & \multirow{2}{*}{KPConv}	& 750(subcloud)  &   &  &    &  &  &  &86.46 &  &74.19  &  &  &   &  &  &   & &   &  &  & &  \\
				& & 1320(subcloud)  &   &  &    &  &  &  &  &  &   &  &  &   &  &  &   & &   &  & 94.61 & &85.20 \\\hdashline
				Dai et al. \cite{10788411} & SPVCNN \cite{tang2020searching}  &  AL   &   &  & 70.41 &  &  &   &  &  &    &  &  &   &  &  &   & &   &  &  & &  \\  \hdashline
				\multirow{2}{*}{DAAL-WS \cite{LEI2024103970}} & \multirow{2}{*}{RandLA-Net}   &  0.015\%   &   &  &  &  &  &   &  &  &    &  &  &   &97.60 &81.91 &   & &   &  &  & &  \\ 
				& &  0.03\%   &   &  &  &  &  &   &  &  &    &  &  &   &  &  &   & &60.36  &   &  & &  \\ 
				\hline
				
				\multirow{2}{*}{Huang et al. \cite{10423773}}  & 				 \multirow{2}{*}{ PointSIFT \cite{jiang2018pointsift}} &  TUM-MLS-2016(FW)  &   &  &  &  &  &   &  &  &    &  &  &   &81.2 &37.5 &45.1  & &   &  &  & &  \\ 
				&  &  DALES(FW)  &80.7  &55.9 & 65.0 &  &  &   &  &  &    &  &  &   &  &  &   & &   &  &  & &  \\ \hdashline
				Thr-MPRNet \cite{10430180}& DGCNN  &   SensatUrban(FW)  &41.40  & 31.58 &  &  &  &   &44.06 &32.41 &    &76.82 &65.25 &   &  &  &   & &   &  &  & &  \\ \hdashline
				Dai et al. \cite{10818527}	& DGCNN &   SensatUrban(FW)  &45.34  & 31.85 &  &  &  &   &48.11 &37.27 &    &82.51 &76.78 &   &  &  &   & &   &  &  & &  \\ \hdashline
				\multirow{5}{*}{Yuan et al. \cite{rs17152618}}	&   \multirow{5}{*}{Point Transformer \cite{zhao2021point} }	&   DALES and Dublin(FW) 0.1\%  &81.8  &  & 65.1 &  &  &   &  &  &    &  &  &   &  &  &   & &   &  &  & &  \\ 
				& & 1\% &83.7  &  & 68.4 &  &  &   &  &  &    &  &  &   &  &  &   & &   &  &  & &  \\ 
				& &  10\%  &85.5  &  &74.0 &  &  &   &  &  &    &  &  &   &  &  &   & &   &  &  & &  \\ 
				& &  100\%    &85.7  &  & 74.7 &  &  &   &  &  &    &  &  &   &  &  &   & &   &  &  & &  \\ \cdashline{3-24}
				
				&  & \multirow{1}{*}{SensatUrban(FW)} &84.2  & 58.1 &  &  &  &   &  &  &    &  &  &   &  &  &   & &   &  &  & &  \\ \hline
				\multirow{2}{*}{Lin et al. \cite{LIN202279}} 	& \multirow{2}{*}{KPConv} &     0.5\%(subcloud)      &80.0  &  & 63.1 &  &  &   &  &  &    &  &  &   &  &  &   & &   &  &  & &  \\ 
				& &     1\%(subcloud)   &81.4  &  & 65.9 &  &  &   &  &  &    &  &  &   &  &  &   & &   &  &  & &  \\ \hline 
				
				WHCN \cite{LU2025129264} 	& KPConv & scene &   &  &  &  &  &   &  &  &    &  &  &   &  &  &   & &   &  & 86.2 &58.3&  \\ \hline
				ADNLC \cite{10677356}        & KPConv  &  noisy lables &78.2  &53.1 &  &80.5 &53.5 &   &  &  &    &  &  &   &  &  &   & &   &  &  & &  \\ \hline
		\end{tabular}}
		\label{tbl:weak_supervisionresult}
		\vspace{-5mm}
	}
\end{table*}

\section{LiDAR Interpretation with Weak Supervision} \label{section_interpretation}

\textcolor{black}{WSL in LiDAR remote sensing can be categorized into three types \cite{zhou2018brief}: \textbf{Incomplete Supervision}, where only a subset of training data is labeled while the rest remains unlabeled; \textbf{Inexact Supervision}, where annotations are coarse-grained, such as providing scene-level labels instead of point-level labels in semantic segmentation; and \textbf{Inaccurate Supervision}, where annotation data may contain errors or noise.
In addition, there is \textbf{domain adaptation (DA)} \cite{8861136}, which utilizes the fully labeled data in the source domain and the small amount or unlabeled data in the target domain to improve the task performance in the target domain by reducing the distribution differences between the two domains. 
\textcolor{black}{Furthermore, domain generalization (DG) aims to learn a model from data in one or more related source domains, enabling it to generalize effectively to any unseen target domain \cite{9847099}.}
Because research on object detection and instance segmentation in LiDAR remote sensing is limited, existing work has focused primarily on semantic segmentation. Therefore, Table \ref{tbl:dataset_comparison} lists commonly used semantic segmentation datasets, while Tables \ref{tbl:weak_supervision} and \ref{tbl:weak_supervisionresult} summarize representative LiDAR remote sensing methods and their accuracies on the evaluated datasets, respectively.
}
\begin{figure}[htbp]
    \centering
    \includegraphics[width=\columnwidth]{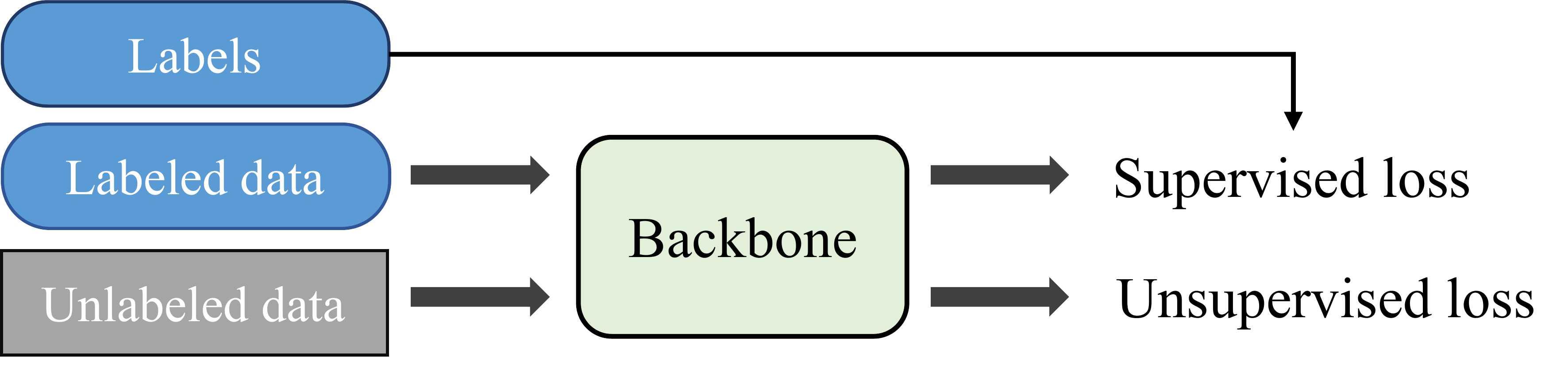}
    \caption{\textcolor{black}{Incomplete supervised learning framework for point clouds.}}
    \label{fig:incompeleSup}
\end{figure}
\subsection{Incomplete supervision}

\textcolor{black}{Incomplete supervision refers to the scenario where, with only limited labeled data providing supervised signals, a large amount of unlabeled data is leveraged to ensure effective network learning~ \cite{zhou2018brief,9751593,10561540}.
Fig.~\ref{fig:incompeleSup} provides a schematic illustration redrawn from \cite{10561540}.
Drawing from the categorization in \cite{zhou2018brief,10561540}, we divide incomplete supervision into four main techniques:
(i) Learning strategies that densely annotate partial regions of point cloud scenes while leaving the rest unlabeled, commonly known as semi-supervised learning~ \cite{zhou2018brief,9751593,10561540};
(ii) Strategies that sparsely annotate only a few points in point cloud scenes, termed WSL with sparse annotations~ \cite{Zhang2021WeaklySS};
(iii) Active learning methods in which the model proactively selects the ``most informative'' samples from unlabeled data for human annotation;
(iv) Few-shot learning that utilizes abundant labeled samples from base classes and limited samples from novel classes to train the model, thereby reducing annotation costs for novel classes.}

\subsubsection{Semi-supervised learning}
It aims to improve network learning performance by utilizing a small portion of fully labeled training data from the scene and a large portion of unlabeled scene data \cite{zhou2018brief}.
Extensive research has been conducted in the point cloud field, such as contrastive learning \cite{jiang_guided_2021}, consistency regularization \cite{9725030}, pseudo-labeling \cite{cheng2021sspcnet}, self-training \cite{li_semi-supervised_2021}, etc.

\textcolor{black}{\textbf{Semi-supervised 3D semantic segmentation} aims to fully leverage all data in a setting where a few scenes are fully labeled while the majority remain unlabeled, in order to achieve comprehensive semantic annotation. 
Li et al. \cite{li_semi-supervised_2021} introduce a GAN-like discriminator within a self-training framework to evaluate pseudo-label confidence. This approach proves more reliable than solely relying on model outputs, but the adversarial training can introduce instability. 
Zhang et al. \cite{9725030} employ the classic Mean Teacher model, utilizing unlabeled data through consistency regularization; however, this class of methods essentially treats a point cloud as a collection of independent points and fails to fully exploit its rich geometric structure priors.
Recognizing the importance of structural information, subsequent research gives rise to methods centered on ``superpoints''. Works by Cheng et al. \cite{cheng2021sspcnet} and Deng et al. \cite{deng_superpoint-guided_2021} first over-segment point clouds into superpoints with local consistency, elevating the segmentation task from the point to the region level to better utilize local geometric and semantic consistency. 
Further research applies semi-supervised methods to LiDAR semantic segmentation in autonomous driving scenarios. 
For example, LaserMix \cite{kong_lasermix_2023} leverages the strong spatial priors of LiDAR data by mixing specific laser beams from different scan frames to generate new samples while maintaining prediction consistency, achieving significant results. However, its heavy reliance on sensor and scene priors limits its generalizability to unstructured indoor scenes or general 3D models.}

\textcolor{black}{In \textbf{LiDAR remote sensing}, most research focuses on ALS point clouds. Early work by Liao et al. \cite{7325695} proposes a semi-supervised graph-fusion framework that combines spectral, spatial, and elevation features from LiDAR and hyperspectral data. With the advent of deep learning, attention shifts to more sophisticated neural architectures. Li et al. \cite{9915611} introduce a ternary deep network (TSDN) that uses dedicated branches to address modality heterogeneity and adopts a multi-model consensus strategy to purify pseudo-labels, thereby enhancing the robustness of semi-supervised learning. Guo et al. \cite{Guo2023SemisupervisedCF} adopt a two-stage ``unsupervised pretraining—supervised fine-tuning'' paradigm, learning general features via reconstruction. Pu et al. \cite{10609507} employ an unsupervised encoder–decoder with data-specific branches for joint training, enabling effective multimodal representations from limited labeled samples. Additionally, Dersch et al. \cite{Dersch2024SemisupervisedMT} apply semi-supervised methods to multi-class tree-crown delineation. More recently, Li et al. \cite{10660507} adopt contrastive learning, replacing pixel-level reconstruction objectives with feature-space constraints that enforce intra-class compactness and inter-class separability to obtain more discriminative representations.
To obtain more discriminative representations. 
Notably, many of these approaches leverage multi-source data fusion, combining LiDAR's geometric strengths with hyperspectral data to overcome issues like \textcolor{black}{same object different spectra (SODS)/same spectrum different objects (SSDO)}. This fusion, while powerful, increases annotation costs, further reinforcing the necessity of semi-supervised techniques to make such advanced analysis practical and scalable.}

\textbf{Semi-supervised 3D object detection} is the task of estimating object classes and 3D bounding boxes for all objects in a scene using labeled and unlabeled data.
It can be divided into two groups: consistency-based methods \cite{NEURIPS2019_d0f4dae8} and pseudo-labeling methods \cite{sohn2020simple}.

Zhao et al. \cite{zhao2021sess} propose a robust 3D object detector trained via the Mean Teacher paradigm, incorporating point cloud-specific perturbations and consistency loss.
Similarly, Wang et al. \cite{9578176} utilize a teacher-student mutual learning framework to propagate information from labeled data to unlabeled data in the form of pseudo-labels.
Outdoor LiDAR point clouds present unique challenges such as complex structures and low point density \cite{9495168}.
Research on 3D point cloud object detection in LiDAR remote sensing is limited.
\textcolor{black}{Current research is largely focused on LiDAR point clouds for autonomous driving \cite{yin2022semi,wu_semi-supervised_2024}.}
For instance,
\textcolor{black}{3DIoUMatch \cite{9578176} introduces pseudo-label filtering based on IoU prediction to eliminate low-quality annotations. However, the reliability of this filtering is compromised in sparse or occluded scenes, where noise in the IoU predictions undermines the selection process. Qi et al. \cite{9578894} leverage full point cloud sequences offline to generate high-quality ``auto-labels'' significantly enhancing label precision. Nevertheless, its offline nature and substantial computational cost render it unsuitable for real-time deployment. 
Most recently, SSF3D \cite{wong2024ssf3d} adopts a ``strict semi-supervised'' strategy using stringent thresholds and filter switching to ensure pseudo-label quality, but this risks biasing the model towards easy samples and thus impairing its generalization ability on more challenging instances.}

\textbf{Semi-supervised 3D instance segmentation} combines a small amount of labeled data with a large amount of unlabeled data to achieve semantic classification and instance differentiation of objects in point clouds, aiming to reduce annotation costs and improve model performance.

Compared to semi-supervised 3D semantic segmentation and object detection, research on semi-supervised 3D instance segmentation is limited \cite{Liao2021PointCI,9879061}.
For instance,
\textcolor{black}{Liao et al. \cite{9506359} leverage consistency regularization—enforcing invariant predictions under input perturbations—to exploit unlabeled data, yet the shallow self-supervised signal struggles to capture the deep correlation between instances and semantics, offering limited improvement when annotations are scarce. Chu et al. \cite{9879061} propose a self-training framework that facilitates mutual reinforcement between semantic and instance pseudo-labels and introduces a learnable ``proposal re-correction module'' to denoise these pseudo-labels at the object level, significantly enhancing pseudo-label quality. However, the two pseudo-label types are prone to error coupling, and the framework's effectiveness depends on the re-correction module's performance. Liu et al. \cite{liu2022weakly}, in contrast, construct regional units via over-segmentation and simultaneously model semantic and instance boundaries using a boundary-aware energy function and contrastive learning. This approach introduces a stronger structural prior, but its performance is constrained by the initial over-segmentation quality, and the joint optimization of the energy function and contrastive objective presents additional challenges.}

\subsubsection{Sparse-supervised learning}
It refers to the scenario in 3D point cloud data where only a small number of points are sparsely labeled, and these limited annotations, combined with large amounts of unlabeled data, are used to train the network effectively. The research methods include consistency regularization \cite{xu_weakly_2020}, contrastive learning \cite{li_hybridcr_2022}, self-training \cite{hu_lidal_2022}, label propagation \cite{zhao_semanticflow_2023}, pseudo-labeling \cite{lei_wspointnet_2022}, active learning \cite{xu_hierarchical_2023}, and others.

\textbf{Sparse-supervised 3D semantic segmentation} refers to performing semantic segmentation on 3D point cloud data where only a subset of points is labeled, while most remain unlabeled. The goal is to train a model capable of accurately segmenting the entire point cloud using limited labeled information.

\textcolor{black}{A large number of well-established works have emerged \cite{xu_weakly_2020,pan_less_2024-1}}.
Xu et al. \cite{xu_weakly_2020} employ a teacher-student model and leverage active learning to select and label the most informative samples, achieving good results with very little labeled data.
Considering information redundancy in existing active learning strategies, Xu et al. \cite{xu_hierarchical_2023} design a feature distance suppression module to select the important points to be labeled.

\textcolor{black}{Hu et al. \cite{hu2022sqn} tackle the challenge of large-scale point cloud segmentation at a minimal cost by querying local semantic information to efficiently propagate sparse supervisory signals, although the core assumption of local semantic homogeneity compromises the segmentation accuracy at object boundaries.}
\textcolor{black}{Some research applies similar strategies like pseudo-labeling, active learning, and self-training to LiDAR point clouds in autonomous driving \cite{hu_lidal_2022,zhao_semanticflow_2023,chen2024weaklysupervisedLiDARsemantic}.}
For instance,
Hu et al. \cite{hu_lidal_2022} propose a new active learning strategy for 3D LiDAR semantic segmentation based on the inconsistency of inter-frame prediction to estimate model uncertainty and further ensemble self-training to improve performance.
\textcolor{black}{Zhao et al. \cite{zhao_semanticflow_2023} bidirectionally propagate and refine pseudo-labels between sparsely annotated full frames to address the high annotation cost in the semantic segmentation of sequential LiDAR point clouds. However, the segmentation performance relies heavily on the accuracy of the scene flow estimation and degrades as the interval between annotated frames increases due to error accumulation.}

\textcolor{black}{Currently, extensive research has been conducted in \textbf{LiDAR remote sensing}, and studies addressing point clouds acquired by different platforms mainly include the following: for ALS point clouds, Wang et al. \cite{isprs-annals-V-2-2021-43-2021} introduce an adaptive-threshold strategy to select reliable pseudo-labels; moreover, they penalize prediction uncertainty with an entropy regularizer and impose consistency between single and fused predictions to improve network robustness \cite{WANG2022237}. Zhang et al. \cite{zhang_perturbed_2021} propose the Perturbed Self-Distillation (PSD) framework, which builds perturbed branches to capture point-cloud graph topology and enforce prediction consistency for information transfer between labeled and unlabeled points; they also explore colorization-based self-supervised pretraining to transfer priors from unlabeled data into weakly supervised networks \cite{Zhang2021WeaklySS}. The Semantic Query Network (SQN) \cite{hu2022sqn} exploits local semantic homogeneity and aggregates multi-scale contextual features (e.g., via interpolation) to implicitly strengthen supervision under extremely sparse annotations (e.g., 0.1\%). Extensions of the teacher–student paradigm, such as HSCN \cite{10735245}, DCUF-Net \cite{10813603} and DSDCL \cite{10855480}, further enhance consistency learning and discriminative feature learning by introducing multi-scale geometric similarity losses, data-and-feature dual perturbations with uncertainty-aware mechanisms, and feature-space pull/push strategies, respectively; DR-Net \cite{10488461} designs a Dilated Region Feature Aggregation (DRFA) module to aggregate local features at multiple dilation scales and enrich representations.}

\textcolor{black}{For MLS point clouds, Guinard et al. \cite{isprs-archives-XLII-1-W1-151-2017} manually select representative points and partition the cloud into geometrically consistent regions, applying a conditional random field (CRF) for context-aware classification to achieve accurate urban scene labeling under severely limited supervision. Lei et al. \cite{lei_wspointnet_2022} propose the multi-branch weakly supervised WSPointNet, which introduces a temporal dimension and maintains an exponential moving average (EMA) of per-point predictions to form a more stable “ensemble teacher” that the consistency objective aligns to, thereby improving robustness. MSCs \cite{10292660} impose scene-level constraints at encoder, decoder and classifier stages to mitigate class imbalance, and Chen et al. \cite{10696960} exploit hand-defined line and plane points as extra supervision to guide feature learning in a line–plane–object hierarchy.}

\textcolor{black}{For multispectral (MS-LiDAR) point clouds, methods aim to integrate multimodal information efficiently and reduce reliance on full supervision: FPWS-Net \cite{chen_feature_2024} uses main/auxiliary semantic inference modules constrained by consistency and mutual pseudo-label losses to enforce agreement on unlabeled data, while SCSQ-Net \cite{10767226} employs a shared kernel point convolution (SKPC) backbone that shares weights across encoding and semantic inference stages to integrate neighborhood information efficiently.
To address long-tailed class imbalance in TLS datasets, Li et al. \cite{LI2024110701} decouple training into two alternating stages: a feature-learning stage that uses resampling or reweighting to force the feature extractor to learn all classes more evenly (especially rare tail classes), and a classifier fine-tuning stage in which the extractor is frozen and the classifier is adjusted using the original biased data, thereby balancing discriminative power and distributional bias. 
Overall, weakly supervised methods across point-cloud types—through targeted mechanisms such as pseudo-label filtering, self-distillation, temporal smoothing, multi-scale consistency, and task decoupling—substantially reduce annotation cost while improving generalization and robustness in real-world scenarios.}

\textbf{Sparse-supervised 3D object detection} refers to detecting objects in 3D point clouds, where only a subset of objects is labeled while most remain unlabeled. The task involves predicting the object categories and their corresponding 3D bounding boxes.
\textcolor{black}{Most research is focused on LiDAR point clouds for autonomous driving, with fewer studies targeting indoor point clouds, while its application in LiDAR remote sensing remains significantly underexplored.}

\textcolor{black}{Meng et al. \cite{9369074} pioneer the use of bird's-eye-view center-point clicks as extremely weak labels to generate cylindrical candidate boxes, which are then optimized with a few precise annotations. Although this approach significantly reduces the annotation burden, it relies on the quality of the initial candidates, exhibits limited generalization to objects with complex or highly variable shapes, and still requires precise annotations.
To compensate for the geometric information missing in weak labels, Xu et al. \cite{9879774} utilize a 3D shape library to construct virtual scenes and transfer knowledge via domain adaptation. However, this method is constrained by the domain gap between virtual and real-world data and the availability of high-quality shape libraries.
For LiDAR point clouds in autonomous driving scenarios, Liu et al. \cite{9880058} discover positive samples through a missing-label instance mining module, design a reliable background mining module, and propose a point cloud completion data augmentation strategy to generate high-confidence data for iterative learning. 
Zhang et al. \cite{10378639} leverage a small number of fully-annotated box labels and a large number of weakly-annotated point labels. By modeling the global interaction between LiDAR points and weak annotations, they generate high-quality pseudo bounding boxes to train any 3D detector. }

\textbf{Sparse-supervised 3D instance segmentation} refers to performing instance-level segmentation on 3D point cloud data where only a subset of points or instances is labeled. The goal is to assign each point to a specific instance ID while predicting the instance's category label.

\textcolor{black}{PointContrast \cite{xie2020pointcontrast} pioneers scene-level unsupervised pre-training using contrastive learning, but its focus on point-level consistency neglects global spatial structure, limiting its performance on complex tasks such as instance segmentation. 
SegGroup \cite{9833393} weakens supervision to a single-point annotation per instance and generates pseudo-labels through segment grouping, which significantly reduces annotation costs but is highly dependent on the initial over-segmentation and grouping strategy, making it prone to error propagation. 
ClickSeg \cite{liu2023clickseg} refines single-point supervision to the ``click-level'', using annotated points as initial seeds for k-means. While this simplifies the pipeline, it remains constrained by the inherent resolution limitations of clustering in dense or occluded scenes. 
Dong et al. \cite{dong2023weakly} advance this line of research further by relying solely on semantic labels and learning an ``objectness'' signal to separate instances. Despite the significance of this step, the core challenge remains how to reliably generate initial instance seeds in regions with clustered same-class objects to avoid erroneous merging or over-segmentation.}

\subsubsection{Active learning}
\textcolor{black}{
Active learning addresses scenarios with abundant unlabeled data and scarce labeled data. It uses an intelligent query process to selectively sample the most informative instances for expert annotation, aiming to maximize model performance at a minimal annotation cost.
Active learning has achieved great success in 2D images.
Its methods can be categorized into three main types: uncertainty-based approaches \cite{9093475}, which select samples with low predictive confidence to maximize information gain and refine decision boundaries; diversity-based approaches \cite{sener2017active}, which prioritize representative and diverse samples to ensure comprehensive coverage of the data distribution and mitigate redundancy inherent in uncertainty methods; and hybrid approaches \cite{siddiqui2020viewal}, which balance informativeness and representativeness, typically by first filtering candidates based on uncertainty and then removing redundancy through diversity criteria (or vice versa).
}

\textcolor{black}{Early works (e.g., ReDAL \cite{9710948} and Shi et al. \cite{shi2021labelefficientpointcloudsemantic}) pioneer shifting the active-learning selection unit from whole scenes to subregions (regions/super-points), using measures such as entropy, color, and structural complexity to identify informative areas while promoting inter-region diversity to reduce redundancy. However, these methods rely on pre-segmentation, which can introduce partitioning bias and leave substantial redundancy inside selected regions. To address this, Annotator \cite{xie2023annotator} proposes a voxel-centric online selection strategy: a “voxel confusion degree” (VCD) that jointly captures uncertainty and diversity, and source-domain pretraining to mitigate cold-start issues, markedly improving labeling efficiency in large-scale scenes. Nearly concurrently, Xu et al. \cite{xu_hierarchical_2023} abandon region-based selection and introduce a hierarchical point-wise scheme that uses a multi-scale uncertainty module (HMMU) to assess point-level value and a feature-distance suppression (FDS) mechanism to ensure representativeness and dispersion—enabling more precise selection under very low annotation budgets. Despite these advances, manual annotation costs and potential annotator bias remain, and most methods focus on segmentation; their extension to other 3D tasks warrants further study.
}

\textcolor{black}{Recent work in \textbf{LiDAR remote sensing} has advanced active-learning methods to reduce annotation cost. Early work by Polewski et al. \cite{7378854} combines active learning and semi-supervised learning within a Rényi-entropy regularization framework: by minimizing the entropy of unlabeled samples, it selects the most informative samples for annotation and substantially improves classification accuracy in early training. Luo et al. \cite{8353504} adopt an active-learning scheme that iteratively labels a small number of unlabeled supervoxels using a neighborhood-consistency prior. With the rise of deep learning for point clouds, Lin et al. \cite{isprs-annals-V-2-2020-243-2020} integrate active learning with deep networks (e.g., PointNet++) and propose a more efficient block-level segmentation entropy for uncertainty estimation. This advance improved feature learning and model performance but revealed a core limitation: uncertainty-based selection tends to pick “difficult” samples that are not necessarily representative, leading to spatially and structurally redundant (low-diversity) selections. To mitigate this, Lin et al. \cite{LIN202073} later introduce an optimized method that combines point entropy, segmentation entropy, and mutual information to guide sample selection, and adopt incremental fine-tuning to reduce annotation needs and training time. Kölle et al. \cite{isprs-annals-V-2-2021-93-2021} combine active learning with crowdsourced annotation, using a Random Forest to select a small set of points from ALS point clouds and textured 3D meshes for efficient coupled semantic segmentation. SSDR-AL \cite{10.1145/3503161.3547820} uses a graph-reasoning network over superpoint diversity to progressively select informative and representative samples and then applies a noise-aware iterative annotation strategy. Kölle et al. \cite{isprs-annals-X-1-W1-2023-945-2023} further shift from a model-centric to a data-centric paradigm and validate active-learning performance. More recent studies by Wang et al. \cite{wang2023one} and Dai et al. \cite{10788411} extend these ideas to extremely sparse supervision (One-Class-One-Click) and propose more efficient uncertainty measures (e.g., ``Temporal Output Discrepancy'') that require no extra model ensembling, as well as human-inspired Spatial Incremental Sampling strategies. These advances substantially improve label efficiency but introduce new challenges—additional hyperparameters and open questions about defining and measuring diversity. Finally, DAAL-WS \cite{LEI2024103970} employs an ensemble-prediction-based active-learning policy (PLEP) that selects points by point-level uncertainty to achieve high-accuracy 3D scene understanding with very few labeled samples.
}

\subsubsection{Few-shot learning}
\textcolor{black}{Few-shot learning aims to rapidly adapt to novel classes and achieve strong generalization when only a very small number of labeled examples are available per class in the target domain. This paradigm has made significant progress in 2D images.
Current mainstream methods typically rely on the meta-learning framework and can be further categorized into three types: metric-based methods \cite{NIPS2017_cb8da676}, which learn an appropriate similarity metric space to perform classification predictions using the similarity (such as cosine similarity or Euclidean distance) between query samples and support set samples; optimization-based methods \cite{Jamal_2019_CVPR}, which optimize the model's initialization parameters or optimization process via meta-learning, enabling rapid adaptation to new tasks with few samples through a limited number of gradient steps; and augmentation-based methods \cite{Chen_2019_CVPR}, which generally enable rapid adaptation through internal model mechanisms or data/memory augmentation.
}

\textcolor{black}{As a pioneering work, AttMPTI \cite{9577428} successfully transplants the prototype-learning paradigm from 2D images to point clouds — using a pretrained encoder to distill class prototypes from few support samples to guide segmentation. However, this direct transfer reveals a fundamental weakness: large scene- and object-shape discrepancies between the support and query sets make prototypes distilled from isolated support examples fragile and unrepresentative in complex query scenes. 
To bridge this gap, both BFG \cite{mao2022bidirectional} and 2CBR \cite{zhu2023cross} explore a ``pretraining + prototype adaptation'' framework, introducing query information to dynamically adjust or “correct” the prototypes. Although these methods yield some performance gains, they do not address deeper limitations such as domain bias introduced by the pretrained encoder and the high training cost; consequently, model efficiency and generalization remain fundamentally constrained. 
A true paradigm shift appears in works represented by TaylorSeg \cite{wang2025taylor}, which abandon costly pretraining in favor of training-free geometric fitting or lightweight learnable modules — greatly improving efficiency and avoiding domain bias at its root, while potentially sacrificing the feature richness afforded by large-scale pretraining. 
Meanwhile, COSeg \cite{10657196} takes a more critical view of the task formulation itself, exposing issues in earlier benchmarks such as “foreground leakage” that fundamentally undermine the reliability of some reported metrics. 
}

\textcolor{black}{In \textbf{LiDAR remote sensing}, few-shot learning remains underexplored. Zhang et al. \cite{isprs-annals-X-4-W5-2024-333-2024}’s Pole-NN achieves a breakthrough in few-shot classification of pole-like objects using a training-free, non-parametric feature extractor, but it is restricted to specific object classes and does not readily scale to complex urban scene segmentation. Huang et al. \cite{10423773} reduce annotation requirements for full-scene segmentation by combining sparse annotations with distribution calibration; however, this pretraining-dependent approach struggles to generalize under large domain shifts (e.g., different sensor types). To tackle cross-domain transfer, Dai et al. \cite{10430180} propose Thr-MPRNet, a meta-learning multi-prototype relation network that transfers knowledge from photogrammetric point clouds to airborne laser scanning (ALS) data, yet meta-learning methods are prone to catastrophic forgetting when novel classes appear continuously in dynamic scenes. Addressing this, Dai et al. \cite{10818527} introduce a cross-domain incremental feature-learning framework that uses a distillation module to dynamically adapt to newly emerging target-domain classes while retaining transferred knowledge. Most prior methods, however, treat geometric, intensity and other cues uniformly and thus ignore modality heterogeneity; Yuan et al. \cite{rs17152618} explicitly point out this shortcoming and propose a multi-branch feature extractor that separates geometric and reflectance features to improve transfer and incremental learning in complex ALS scenarios. Together, these developments signal a shift from purely algorithmic strategies toward more fine-grained feature-representation learning.
}

\subsubsection{Discussions for Incomplete supervision}

\textcolor{black}{Unlike optical imagery, LiDAR point clouds are inherently sparse, irregularly sampled, and spatially discontinuous, especially in regions affected by occlusion or variable scanning patterns. Such modality characteristics exacerbate incomplete supervision, where labeled points cover only a small fraction of the 3D scene and exhibit strong spatial and structural imbalance.
The core strategy of weakly supervised learning under incomplete labels is to propagate limited annotations through pseudo-labeling and consistency regularization, enabling the model to leverage the abundant unlabeled data. However, applying these methods directly to LiDAR faces modality-specific challenges: sparse and uneven sampling interrupts label propagation, vertical structural layering leads to semantic misalignment between canopy and ground points, and noise distribution varies drastically across scanning patterns.}

\textcolor{black}{To address these issues, LiDAR-specific adaptations primarily introduce structure-aware pseudo-label generation and geometry-constrained consistency learning. Specifically, label propagation relies not only on feature similarity but also on geometric primitives (e.g., planarity, linear structures, and surface normals) for guidance, ensuring that pseudo-labels conform to physical structures. Multi-scale neighborhood queries are employed to better exploit the local and global geometry of point clouds, while customized data augmentation strategies enable learning representations robust to point sparsity, scanning variations, and noise.
These targeted strategies mitigate the inherent sparsity, structural complexity, and noise of LiDAR, enabling weakly supervised methods to reliably amplify sparse annotations, narrow the supervision gap, and preserve geometric fidelity in large-scale 3D environments.
}

\subsection{Inexact supervision}

Inexact supervision refers to providing coarse-grained labels instead of precise point-wise annotations. The goal is to train a model using rough labeling information to achieve accurate point-level tasks. It typically includes sub-cloud-level annotations \cite{LIN202279}, scribble annotations \cite{Unal2022ScribbleSupervisedLS}, bounding box annotations \cite{lu2024bsnet}, and scene-level annotations \cite{xia2023densify}, among others.

\begin{figure}[htbp]
    \centering
    \includegraphics[width=\columnwidth]{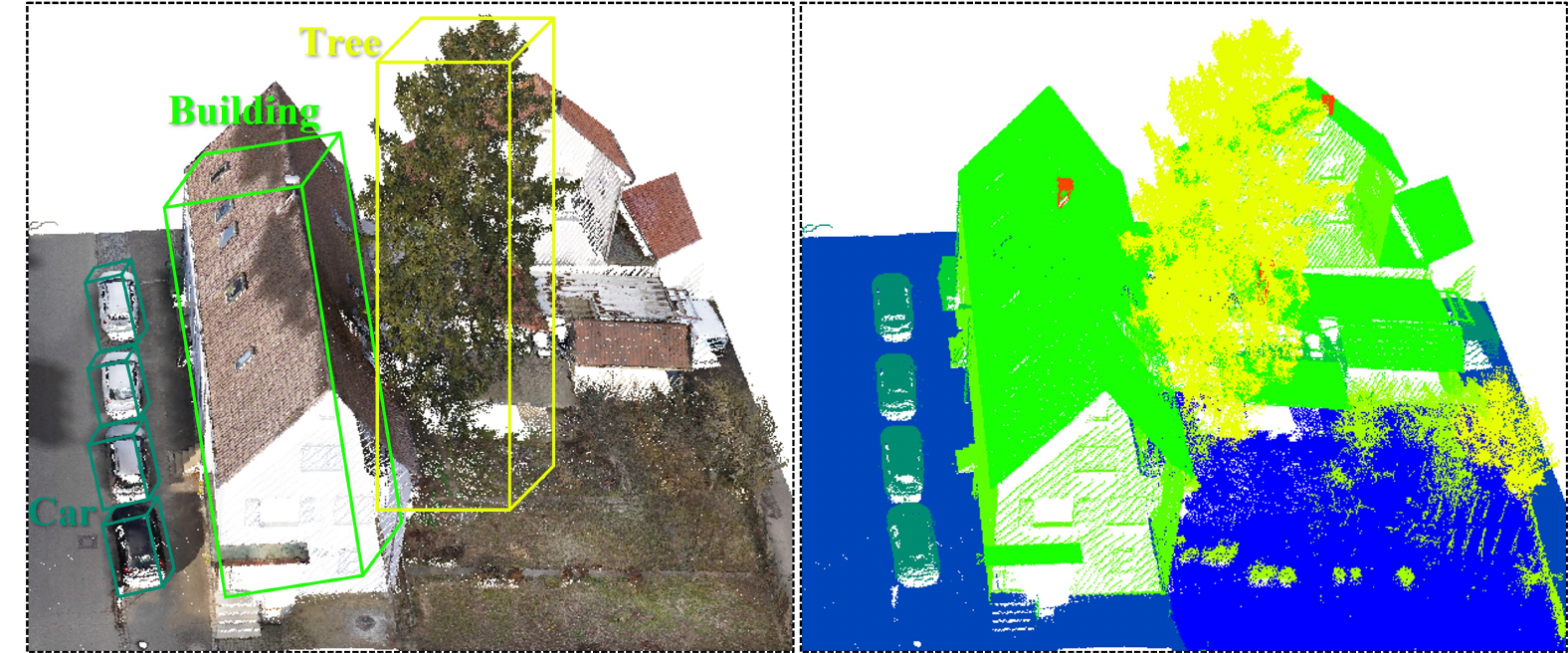}
    \caption{Bounding box annotations (left) and point-wise semantic labels (right) from the H3D dataset \cite{KOLLE2021100001}.}
    \label{fig:box}
\end{figure}
 \begin{figure}[htbp]
    \centering
    \includegraphics[width=\columnwidth]{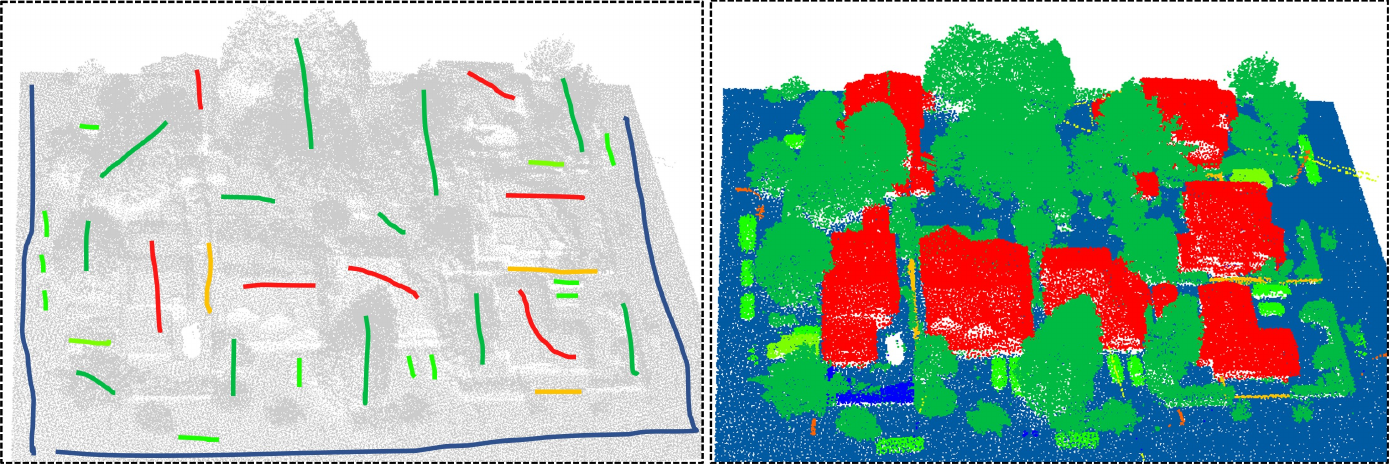}
    \caption{Illustration of scribble-labeled LiDAR point cloud scenes (left) and corresponding overlaid frames (right) from the DALES dataset \cite{9150622}.}
    \label{fig:scribble}
\end{figure}
\begin{figure}[htbp]
    \centering
    \includegraphics[scale = 0.14]{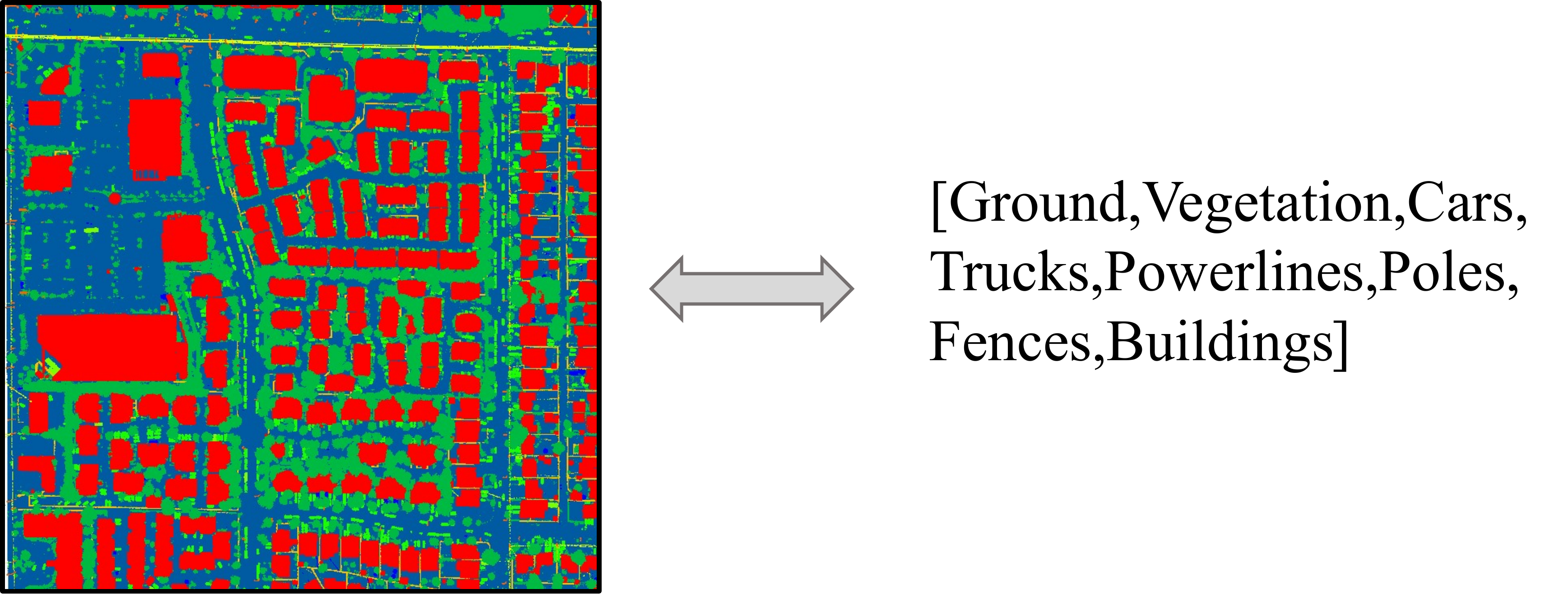}
    \caption{Example of scene-level annotated point cloud from the DALES dataset \cite{9150622}.}
    \label{fig:scene}
\end{figure}

 
\textbf{3D semantic segmentation}.
Liu et al. \cite{liu_box2seg_2022} utilize \textit{box-level annotations}, as shown in Fig. \ref{fig:box}, and combine them with self-training and point activation maps to reduce the cost of point-wise labeling.
Unal et al. \cite{Unal2022ScribbleSupervisedLS} use \textit{scribbles} to annotate LiDAR point clouds, as shown in Fig. \ref{fig:scribble}, which also significantly improves the efficiency of point cloud annotation. 
\textcolor{black}{Some studies use \textit{subcloud-level} labels to reduce annotation costs, assigning the classes present within the sampled regions as labels \cite{wei_multi-path_2020,LIN202279}.}
For instance, Wei et al. \cite{wei_multi-path_2020} propose a point cloud scene segmentation network from cloud-level weak labels on indoor 3D data, and they introduce a multi-path region mining module.
Compared to the above labeling methods, more studies prefer using \textit{scene-level} labels \cite{ren20213d,Yang_2022_CVPR,xia2023densify}, where only the class information of the entire scene is provided for 3D point clouds, without annotating individual points or object instances, as shown in Fig. \ref{fig:scene}.
For instance,
\textcolor{black}{Ren et al. \cite{ren20213d} extract supervision from scene-level labels within a multi-task framework using multiple-instance learning (MIL) and cross-task consistency losses, demonstrating the feasibility of 3D recognition with only scene labels, but applying scene labels uniformly to all points produces sparse, ambiguous supervision, noisy pseudo-labels, and imprecise boundaries that limit segmentation accuracy.
To address this, Xia et al. \cite{xia2023densify} discover semantic “primitives” in the point-cloud feature space via unsupervised clustering and conservatively assign scene labels to related primitives using bipartite matching, converting global label broadcasting into local associations, thereby densifying supervision and significantly improving pseudo-label quality and segmentation performance.
}

 \begin{figure}[htbp]
    \centering
    \includegraphics[width=\columnwidth]{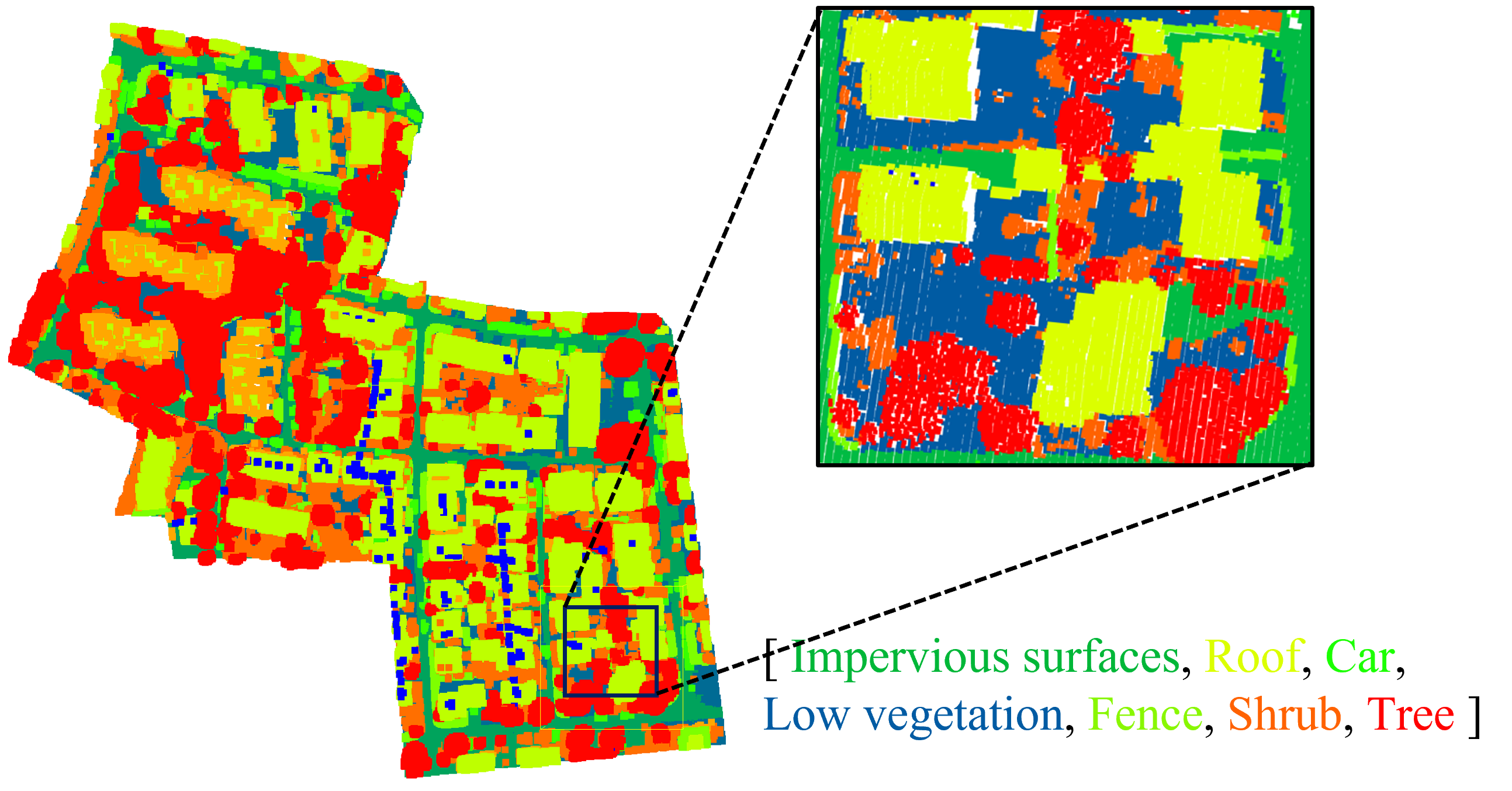}
    \caption{Example of subcloud-level annotated point cloud from the ISPRS dataset \cite{isprs-dataset}.}
    \label{fig:subcloud}
\end{figure}
In \textbf{LiDAR remote sensing},
\textcolor{black}{
Lin et al. \cite{LIN202279} propose a weakly supervised semantic segmentation method for \textbf{ALS point clouds}. They leverage weak subcloud annotations, as shown in Fig. \ref{fig:subcloud}, first generating accurate pseudo-labels using a classification network with overlap region loss and elevation attention, and then optimizing a segmentation network with supervised contrastive loss to efficiently learn from limited subcloud labels.
Wang et al. \cite{wang2023one} train by assigning a point to each class label of the sub-cloud, incorporating contextual constraints and pseudo-label generation. They use active learning based on Temporal Output Discrepancy (TOD) to select the most informative sub-point clouds for labeling, achieving near fully-supervised segmentation performance with minimal labeled data on ALS, MLS, and TLS datasets.
WHCN \cite{LU2025129264} constructs a hypergraph from high-confidence superpoints derived from scene-level annotations. This hypergraph is then used as input to predict high-precision point-level pseudo-labels, which are subsequently used to train a segmentation network.}

\textbf{3D object detection}.
\textcolor{black}{Some studies use \textit{position-level} annotations instead of 3D bounding boxes for 3D object detection to reduce annotation costs \cite{meng2020weakly,9879774}.}
For instance,
Meng et al. \cite{meng2020weakly} combine bird's-eye view (BEV) center-click annotations with a small number of precise 3D instance labels to significantly reduce annotation costs.
Using Scene-level annotation is a simpler approach.
WyPR \cite{ren20213d} uses simple scene-level labels and addresses the problem of weakly supervised 3D point cloud object detection through multi-task learning and self-supervised constraints.
Qin et al. \cite{qin2020weakly} combine scene-level labels with results from a pre-trained network applied to images mapped from 3D point clouds as soft labels, significantly reducing annotation costs while maintaining 3D object detection performance.

\textbf{3D instance segmentation}.
\textcolor{black}{Some research uses \textit{box annotations} to reduce labeling costs \cite{ngo2023gapro,chibane2022box2mask,lu2024bsnet}.
For instance,
Box2Mask \cite{chibane2022box2mask} first demonstrates that dense segmentation models can be trained using only bounding boxes: it assigns points to boxes by voting and obtains segmentations via non-maximum suppression, but relies on simple heuristics for overlapping or nested boxes (e.g., assigning points to the smallest-volume box), which induces misassignments and limits the performance ceiling. To address this limitation, GaPro \cite{ngo2023gapro} proposes a two-stage paradigm: it first generates high-quality pseudo-labels from bounding boxes and then trains arbitrary segmentation networks on them. GaPro models point membership in overlapping regions probabilistically with Gaussian processes and provides uncertainty estimates, offering greater flexibility and improved performance compared with hard assignments, but it remains constrained by the accuracy of the bounding boxes.}

\subsubsection{Discussions for inexact supervision}

\textcolor{black}{Unlike incomplete supervision, where the primary difficulty lies in propagating sparse labels, inexact supervision faces challenges unique to LiDAR’s modality: coarse labels often span multiple objects or vertical layers, leading to severe label leakage; the geometric boundaries in 3D space are blurred by scene- or subcloud-level supervision; and occlusion effects combined with platform-dependent scanning patterns induce pseudo-label drift. 
\textcolor{black}{To address these issues, recent methods introduce a dedicated elevation-attention module to enhance critical height information in airborne LiDAR, and employ superpoint generation together with hypergraph convolutional networks to model complex higher-order structural relationships among objects in outdoor scenes.}
These strategies make weak supervision more aligned with LiDAR’s structural characteristics, enabling more precise segmentation and recognition under coarse labeling conditions.}

\begin{figure}[H]
    \centering
    \includegraphics[width=\columnwidth]{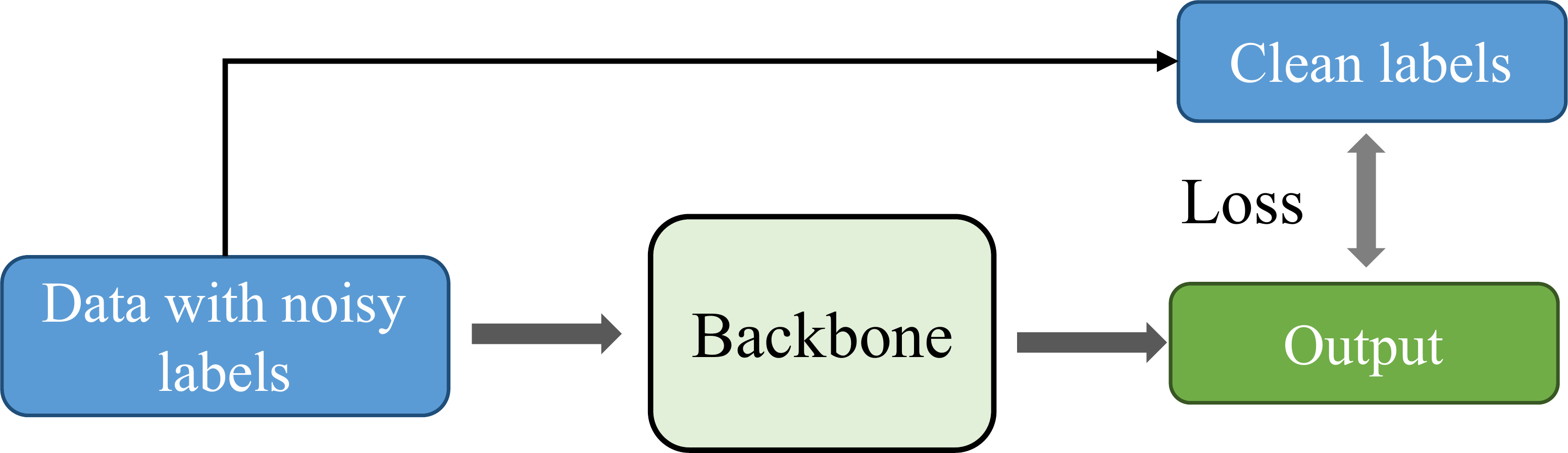}
    \caption{Inaccurate supervised learning framework for point clouds.}
    \label{fig:inaccurateSup}
\end{figure}

\subsection{Inaccurate supervision}

Inaccurate supervision refers to cases where labels contain errors or noise, causing the supervisory signal to deviate from the true information, as shown in Fig. \ref{fig:inaccurateSup}. Techniques like sample selection \cite{Han2018CoteachingRT} and label correction \cite{Huang2020SelfAdaptiveTB} help mitigate the negative impact of inaccurate labels on model training and improve performance.

Research on robust learning with noisy labels in point clouds is limited.
Ye et al. \cite{ye_learning_2021} propose the novel Point Noise Adaptive Learning (PNAL) framework, which uses point-by-point confidence of historical predictions for each point to select reliable labels and considers neighborhood correlation to generate the best labels using a voting strategy. 
Chen et al. \cite{chen2023learning} consider the paradigm of teacher-student models and design a noise-resistant instance supervision module to eliminate the negative effects of noisy labels from a noise learning perspective.
Gao et al. \cite{10677356} propose a new semantic segmentation framework for handling \textbf{ALS point cloud} data with noisy labels, which adaptively corrects noisy labels based on the learning state of different categories to enhance network robustness against noise.

\textcolor{black}{In recent years, the significant progress of 2D foundation models, such as CLIP \cite{Radford2021LearningTV} and SAM \cite{Kirillov2023SegmentA}, has provided new opportunities for weakly supervised learning with inaccurate labels. A common approach is to first perform 2D segmentation on the projected images of a point cloud and then maps the segmentation results back to the point cloud through methods like re-projection to generate pseudo-labels, which are used as imprecise annotations \cite{huang2023segment3dlearningfinegrainedclassagnostic,dong2023leveraginglargescalepretrainedvision,timoneda2024multimodalnerfselfsupervisionLiDAR}. For example, Segment3D \cite{huang2023segment3dlearningfinegrainedclassagnostic} utilizes high-quality segmentation masks generated by SAM to automatically label point clouds as a substitute for manual annotations. However, its performance ceiling and generalization capability are entirely constrained by SAM and the domain of the training data.
}

\subsubsection{Discussions for inaccurate supervision}
\textcolor{black}{Unlike inexact supervision, where the core difficulty stems from coarse and imprecise annotations, inaccurate supervision in LiDAR exhibits modality-specific failure modes. First, noisy labels tend to propagate structurally through sparse yet extended geometric neighborhoods, amplifying their negative impact. Second, when pseudo-labels are derived from 2D foundation models, cross-modal projection errors and viewpoint inconsistencies exacerbate label noise, especially near object boundaries. Third, LiDAR’s sparse and incomplete geometric structures make it difficult to correct such noise, leading to error accumulation and fragile supervisory signals. Recent works address these challenges through confidence-driven filtering, multi-view voting, and noise-adaptive correction, improving robustness against label noise in large-scale LiDAR segmentation.}

\begin{table}[hbt!]
\centering
\caption{\textcolor{black}{Domain adaptation settings.}}
{
\tabcolsep2pt
\resizebox{\columnwidth}{!}{
\begin{tabular}{l|cc|cc}
\hline
\multirow{2}{*}{\centering Setting}                  & \multicolumn{2}{c|}{Data} & \multicolumn{2}{c}{Learning}  \\ \cline{2-5} 
                                           & Source  & Target          & Train stage      & Test stage \\ \hline
Domain adaptation                                    &  $x^s, y^s$     & $x^t, y^t$      & Yes              & No         \\
Unsupervised domain adaptation                          &  $x^s, y^s$    &  $x^t$      & Yes              & No         \\
Source free domain adaptation                                 & -     & $x^t$      & Yes    & No       \\
Test-time adaptation                        & -     & $x^t$      & No (pre-trained) & Yes        \\ 
\multicolumn{1}{l|}{Continual test-time adaptation} & -       & $x_{1}^t, x_{2}^t,......, x_{n}^t$  & No (pre-trained) & Yes        \\ \hline
\end{tabular}
}}
\label{tbl:domain adaptation}
\end{table}
\subsection{Domain adaptation}

The methods described above are primarily designed for a single domain, where applying a pre-trained model to the same domain achieves optimal results. However, remote sensing data is highly affected by temporal, seasonal, and regional variations, leading to domain shift in the acquired LiDAR data. This issue is further amplified by the use of different sensor platforms (e.g., ALS, TLS, MLS), causing a performance decline when pre-trained models are applied to shifted domains.

To address this issue, \textcolor{black}{``\textbf{domain adaptation (DA)}''} aims to transfer knowledge from a labeled source domain to an unlabeled target domain, thereby reducing annotation costs and mitigating the effects of domain shift. It can be regarded as an effective WSL approach, as it does not require labeled data from the target domain, offering a practical solution to the challenges of expensive and time-intensive annotation.
It includes unsupervised domain adaptation, source-free domain adaptation, (continual) test-time adaptation, and so on. Their differences are shown in Table \ref{tbl:domain adaptation}.

\subsubsection{Unsupervised domain adaptation}
Unsupervised Domain Adaptation (UDA) aims to leverage a labeled source domain model to improve performance on an unlabeled target domain. The main challenge is to transfer domain-invariant features from the source to the target domain. Most UDA research has focused on 2D images \cite{He2022NoiseresidualMF}, with only a few studies addressing 3D point clouds. Several approaches have been proposed to tackle the UDA problem for 3D point clouds, including adversarial learning, self-training, feature alignment, and so on.

Pointdan \cite{qin2019pointdanmultiscale3ddomain} is the first work to address point cloud classification in the UDA context, which leverages local and global feature modeling to extract multi-scale information and employs adversarial training to minimize the distribution discrepancy between the source and target domains.
\textcolor{black}{Following Pointdan, these studies also adopt the adversarial learning strategy to solve the 3D UDA problem \cite{Zhao_Wang_Li_Wu_Gao_Xu_Darrell_Keutzer_2020,Li_2023_CVPR}.}
Achituve et al. \cite{Achituve2020SelfSupervisedLF} propose a novel approach for DA on point clouds by combining the \textit{self-supervised} task of Deformation Reconstruction (DefRec) with Point Cloud Mixup (PCM). DefRec simulates local deformations of point clouds and reconstructs the original shapes, enabling the model to learn cross-domain geometric and semantic features. PCM enhances generalization by generating new training samples through random sampling and mixing of point clouds.
\textcolor{black}{Others also apply self-supervised learning to UDA tasks \cite{9412483,Grill2020BootstrapYO}.}

\textit{Self-training} is another line for this task, which leverages pseudo-labels to learn target knowledge gradually.
Cardace et al. \cite{9665900} propose the RefRec method, which leverages shape reconstruction tasks to learn global shape descriptors for refining pseudo-labels. Combined with self-training strategies and domain-specific classification heads, the method effectively mitigates domain discrepancies, enhancing the performance of UDA for point cloud classification.
\textcolor{black}{Some methods based on self-training mechanisms have achieved SOTA performance in 3D UDA \cite{Hu2023DensityInsensitiveUD,Peng2022CL3DUD}.}

\textcolor{black}{Additionally, some methods adopt \textit{feature alignment} strategies \cite{Saltori2020SFUDA3DSU,wang2024syn}.}
Wang et al. \cite{wang2024syn} address the complex domain discrepancies in indoor synthetic-to-real 3D object detection using object-aware augmentation, hierarchical domain alignment (class-level and holistic-level), and pseudo-label optimization strategies.
\textcolor{black}{
To address failures of domain adaptation caused by large distribution differences introduced by different LiDAR sensors (e.g., varying beam counts or scanning patterns), SALUDA \cite{10550726} introduces a self-supervised implicit surface reconstruction auxiliary task that forces the model to reconstruct scene geometry on both source and target data, thereby learning shared geometric features. It strongly depends on geometric information for domain alignment and may be limited when domain gaps are dominated by non-geometric factors (e.g., differing noise patterns).}

\textcolor{black}{In recent years, with the rapid development of \textit{foundation models}, some studies have integrated them to address point cloud UDA  tasks\cite{Peng2023LearningTA,Wu2024CLIP2UDAMF,Xu2024VisualFM}.}
Peng et al. \cite{Peng2023LearningTA} leverage the 2D image embeddings provided by the Segment Anything Model (SAM) as a bridge to align 3D point cloud features from different domains into a unified feature space. 
Wu et al. \cite{Wu2024CLIP2UDAMF} utilize the vision-language knowledge of the CLIP model to generate task-specific text embeddings through Visual-driven Prompt Adaptation (VisPA) and achieve multi-modal feature fusion via Text-guided Context Interaction (TexCI).
Xu et al. \cite{Xu2024VisualFM} leverage the knowledge priors of Visual Foundation Models (VFMs) to generate high-precision pseudo labels (VFM-PL) and employ a FrustumMixing strategy to semantically mix source and target domain data, reducing distribution discrepancies. 

However, research on point cloud UDA in the \textbf{LiDAR remote sensing} domain remains limited.
\textcolor{black}{Some methods apply the same strategy \cite{Luo2020UnsupervisedSA,peng2020point,Xie2021EXPLORINGCS,10179997,Wang2024ADA,10586968}: feature alignment.}
For instance,
Luo et al. \cite{Luo2020UnsupervisedSA} effectively improve cross-scene MLS point cloud semantic segmentation performance through adaptive data and feature alignment on the target domain point cloud.
Peng et al. \cite{peng2020point} align local geometric features and global features and reduce the distribution discrepancy between source and target domains through adversarial training.
Xie et al. \cite{Xie2021EXPLORINGCS} enhance the model generalization through data augmentation techniques and introduce the Deep CORAL method, which aligns second-order statistics of features to achieve cross-city semantic segmentation of ALS point clouds.
Shen et al. \cite{10179997} alleviate the data domain gap by selecting similar scenes based on the scene height distribution, and align the feature distributions using the adversarial learning module (ALM) that solves the DA problem from  TLS data to MLS data.
\textcolor{black}{Wang et al. \cite{10641953} integrate a simple geometric transformation data augmentation technique to enable cross-modal learning from LiDAR point clouds to synthetic aperture radar point clouds.
PS-UDA \cite{10586968} employs an L2-paradigm constraint and a Laplace matrix to pre-align cross-scene features.
Luo et al. \cite{LUO2025537}consider reconstructing the continuous potential scene surface of MLS point clouds and transform both the source and target domain MLS point clouds into a canonical domain to achieve point-level distribution alignment.
}

\subsubsection{Source-free domain adaptation}
Traditional DA requires simultaneous access to both source domain data and target domain data for model adaptation. \textcolor{black}{In contrast, source-free domain adaptation (SFDA) relies solely on a pre-trained model from the source domain and performs adaptation to the target domain without access to source domain data \cite{liang2020we}.} This is achieved by leveraging unlabeled target domain data and utilizing pseudo-labels or feature distributions generated by the model to update its parameters for improved performance in the target domain.

Currently, SFDA is predominantly focused on 2D image processing, while research in 3D point clouds remains relatively limited.
\textcolor{black}{Some research applies pseudo-labeling to solve SFDA \cite{Saltori2020SFUDA3DSU,Saltori2022GIPSOGI,Wang2024MultiConfidenceGS}.}
For instance,
Saltori et al. \cite{Saltori2020SFUDA3DSU} propose the first SFDA framework for LiDAR-based 3D object detection.
Wang et al. \cite{Wang2024MultiConfidenceGS} employ pseudo-label optimization and prototype matching to achieve point cloud primitive segmentation under the SFDA paradigm.
\textcolor{black}{Michele et al. \cite{michele2024train} introduce a simple reference model and use prediction consistency between it and the model under training as an automatic stopping criterion to mitigate performance degradation caused by training instability during target-domain adaptation. However, its effectiveness partly depends on the quality of the chosen reference model, and its training loss assumes similar class distributions in the source and target domains—an assumption that can break down under large domain shifts.}

\subsubsection{Test-time adaptation}

Test-Time Adaptation (TTA) refers to the process of dynamically optimizing a model during the testing phase using unlabeled target domain data to address domain shift, without accessing source domain data or ground-truth labels. Continual Test-Time Adaptation (CTTA) extends TTA by enabling the model to progressively adapt to continuously changing target domains while retaining its adaptability to previous data distributions, making it suitable for more complex and dynamic scenarios.
Most (C)TTA methods are currently applied to 2D images, including techniques such as batch normalization calibration \cite{Mirza2021TheNM}, entropy minimization \cite{Wang2021TentFT}, contrastive learning \cite{Dbler2022RobustMT}, and pseudo-labeling \cite{Chen2022ContrastiveTA}. 

\textcolor{black}{Research on TTA for 3D point clouds remains relatively limited.
The existing methods are mainly based on pseudo-labeling combined with other techniques  \cite{zou2025hgl}.
For instance,
GIPSO \cite{Cao2023MultiModalCT} pioneers the ``source-free online unsupervised domain adaptation (SF-OUDA)'' scenario: given only a pretrained model and no source-domain data, it performs online adaptation via geometric feature propagation and self-training, without access to source data or target labels. However, because it relies solely on the LiDAR modality, its performance degrades significantly when LiDAR becomes unreliable (e.g., due to rain, fog, or sensor noise), and its adaptation process lacks explicit mechanisms to prevent catastrophic forgetting, making it hard to balance adapting to new distributions while retaining source-domain knowledge in continuously changing environments.
To address these limitations, CoMAC \cite{Saltori2022GIPSOGI} extends the problem to ``multi-modal continuous test-time adaptation (MM-CTTA)''. It adaptively shifts reliance to a more reliable modality (e.g., images) when LiDAR quality drops, improving segmentation robustness, and introduces ``Class-Wise Momentum Queues'' that randomly restore pseudo-source features to “rehearse” previously learned knowledge, thereby mitigating catastrophic forgetting and achieving more stable performance in dynamic environments.}
Research on TTA is also included in other fields such as point cloud classification \cite{Shim2024CloudFixerTA}, registration \cite{Jiang2024PCoTTACT}, super-resolution \cite{Hatem2023TestTimeAF}, and so on.

\textcolor{black}{For \textbf{LiDAR remote sensing},
Wang et al. \cite{Wang2024TesttimeAF} are the first to introduce test-time adaptation (TTA) to the geospatial point cloud domain: they progressively update batch-normalization (BN) statistics and fine-tune BN with self-supervised objectives such as information maximization and pseudo-labeling, enabling a pretrained model to adapt to target domains that differ substantially from the source. However, their method follows the conventional TTA paradigm—adapting once to a single static target domain—and does not adequately address catastrophic forgetting in continuously changing data streams.
To address this, APCoTTA \cite{gao2025apcottacontinualtesttimeadaptation} advances to continuous test-time adaptation (CTTA) and proposes ALS-specific solutions: it uses gradient signals to dynamically select and update “low-confidence” layers while freezing “high-confidence” layers that retain source knowledge; it applies entropy-based confidence filtering to remove unstable samples and reduce erroneous pseudo-labels; and it introduces stochastic parameter interpolation to fuse some updated parameters with the original source parameters. These designs balance adapting to new distributions with preserving prior knowledge and substantially mitigate forgetting and error accumulation. Nonetheless, APCoTTA’s robustness under extreme class imbalance or when encountering novel open-set classes remains to be explored.
}

\subsubsection{Discussions for domain adaption}

\textcolor{black}{Unlike inexact or noisy supervision, domain adaptation in LiDAR faces modality-specific challenges arising from spatiotemporal variability and sensor heterogeneity rather than label issues. First, non-stationary spatiotemporal shifts driven by temporal, seasonal, and regional factors lead to dynamic domain drift. Second, cross-platform geometric discrepancies between ALS, TLS, and MLS cause structural distribution breaks, making feature alignment substantially harder. Third, noise pattern and coverage mismatches amplify alignment instability.  
\textcolor{black}{To address these LiDAR-specific challenges, recent domain-adaptation methods integrate data augmentation, multi-level feature alignment, dynamically updated batch normalization (BN) layers, and other adaptation strategies.}
Unlike conventional image DA, which mainly deals with static, texture-driven style shifts, LiDAR DA must cope with non-stationary spatiotemporal drift, cross-platform geometric discrepancies, and sparse, irregular sampling, requiring fundamentally different alignment strategies beyond conventional 2D paradigms. However, current research remains relatively limited in scope, with most efforts focusing on cross-dataset adaptation, leaving broader challenges such as temporal domain shift, cross-platform generalization, and multimodal adaptation largely underexplored.}

\subsection{Domain generalization}

\textcolor{black}{Domain generalization (DG) aims to train a model solely on source domain data, enabling it to perform well on unseen target domains—such as new sensors, locations, or weather conditions—without additional training or fine-tuning. Current research is predominantly focused on autonomous driving, which is critical for safety-critical systems.
Geometry-driven methods, such as those proposed by Sanchez et al., mitigate sensor discrepancies by constructing computationally intensive ``pseudo-dense'' point clouds \cite{sanchez2023domain}.
However, this approach shifts the generalization burden to the localization module at the cost of real-time performance. Kim et al. address this by simulating sparsity during training or by treating a single source domain as a collection of multiple densities to learn density-invariant features \cite{kim2023single}.
This approach is limited by its excessive focus on density, potentially overlooking differences in object shapes and scene layouts.
Li et al. \cite{li2024domain} combine density resampling with test-time adaptation to enhance generalization, albeit with a partial loss of inference speed. In a different approach, Lehner et al. \cite{lehner20223d} employ adversarial learning to generate physically plausible object deformations, thereby improving robustness against anomalous shapes, such as vehicles in accidents.
Furthermore, a large-scale empirical study by Eskandar \cite{eskandar2024empirical} demonstrates that fundamental design choices, such as adjusting anchor box sizes or selecting appropriate training data (e.g., using clear-weather data to generalize to adverse conditions), also significantly impact generalization performance.
In contrast, research in the field of \textbf{LiDAR remote sensing} on domain generalization remains relatively scarce.}

\subsection{Summary and discussion}

\textcolor{black}{Unlike structured and dense 2D imagery, LiDAR point clouds arise from active laser scanning, producing inherently sparse, irregular, and non-uniform measurements. This violates the dense spatial continuity underlying many weak supervision techniques such as label propagation or consistency regularization. Sparse far-field regions disrupt neighborhood-based propagation, leading to unstable supervision and error accumulation. Multi-echo data further entangle vertical semantics (e.g., canopy–understory–ground) and induce heterogeneous noise distributions, breaking the one-to-one label–instance assumption. Severe occlusions create irreversible “data voids” beyond what self-training can recover. In addition, non-stationary spatiotemporal variability introduces domain shifts and unstable noise fields, undermining stationary distribution assumptions widely adopted in WSL frameworks.}

\textcolor{black}{To address these modality-specific challenges, recent WSL methods integrate geometry- and noise-aware strategies: pseudo-labeling and multi-modal fusion compensate for supervision gaps; elevation-aware modules and geometric priors effectively mitigate semantic ambiguity in the vertical dimension; redesigned consistency regularization accommodates irregular sampling; shape priors handle occlusion-induced voids; and domain/test-time adaptation aligns features under platform shifts. However, most studies still focus on static cross-dataset settings and offer limited solutions to occlusion heterogeneity and dynamic spatiotemporal shifts—core characteristics of LiDAR remote sensing. Bridging this gap requires rethinking WSL as a semantic- and geometry-aware framework rather than a direct adaptation of image-based strategies.}

\textcolor{black}{In summary, WSL leverages weak signals through pseudo-labeling, multi-modal fusion, consistency regularization, and structural priors to alleviate LiDAR’s intrinsic challenges, with notable success in semantic segmentation. Yet progress on more complex tasks (e.g., detection, instance segmentation) remains limited, and key modality-specific issues such as waveform complexity, occlusion heterogeneity, and temporal drift are underexplored. The emergence of Foundation Models offers a promising path to address these gaps by generating high-quality labels at scale and reducing annotation costs.}

\begin{table*}[htbp!]
	\centering
	\renewcommand\arraystretch{1.0}
	\caption{\textcolor{black}{Open LiDAR-based inversion datasets. AGB denotes aboveground biomass, and CH denotes canopy height.}}
	\vspace{-3mm}
	\footnotesize
	\setlength{\tabcolsep}{3.5pt}{
		\resizebox{\textwidth}{!}{
			\begin{tabular}{c|c|c|cc|cc|cc} 
				\hline
				\multirow{2}{*}{\textbf{Dataset}} & \multirow{2}{*}{\textbf{Purpose}}& \multirow{2}{*}{\textbf{Time}} & \multicolumn{2}{c|}{\textbf{Extent}}  & \multicolumn{2}{c|}{\textbf{Images}}  & \multicolumn{2}{c}{\textbf{Ground Truth}}  \\ 
				&       &                               & Scope & Surface     & Sensor & Res.     & Sensor & Res.   \\  \hline 
				BioMassters \cite{nascetti2023biomassters}  	& \multirow{2}{*}{AGB}  &  2023  &  Finland  & 85$\times10^3$ km$^2$ & Sentinel-1/Sentinel-2 & 10m  & ALS & 5 points/m$^2$   \\ \cdashline{3-9}
				AGBD \cite{sialelli2024agbdglobalscalebiomassdataset}  	&   &  2025  &  Global (11 regions)  & 5000$\times10^3$ km$^2$ & Sentinel-2/PALSAR-2/ALOS World 3D/Copernicus Land Cover/Canopy Height & 10m  & GEDI & 25m   \\ \hline 
				Lang \cite{lang2023high}	& \multirow{2}{*}{CH}  &  2023  &  Global  & 14k$\times10^3$ km$^2$ & Sentinel-2 & 10m  & GEDI & 25m   \\ \cdashline{3-9}
				Open-Canopy \cite{11095116}	&   &  2025  &  France  & 87$\times10^3$ km$^2$ & SPOT 6-7 & 1.5m  & ALS & 10 points/m$^2$   \\ 
				\hline
		\end{tabular}}
		\label{tbl:reversion_dataset}
		\vspace{-5mm}
	}
\end{table*}

\begin{figure}[htbp] 
    \centering
    \includegraphics[scale = 0.05]{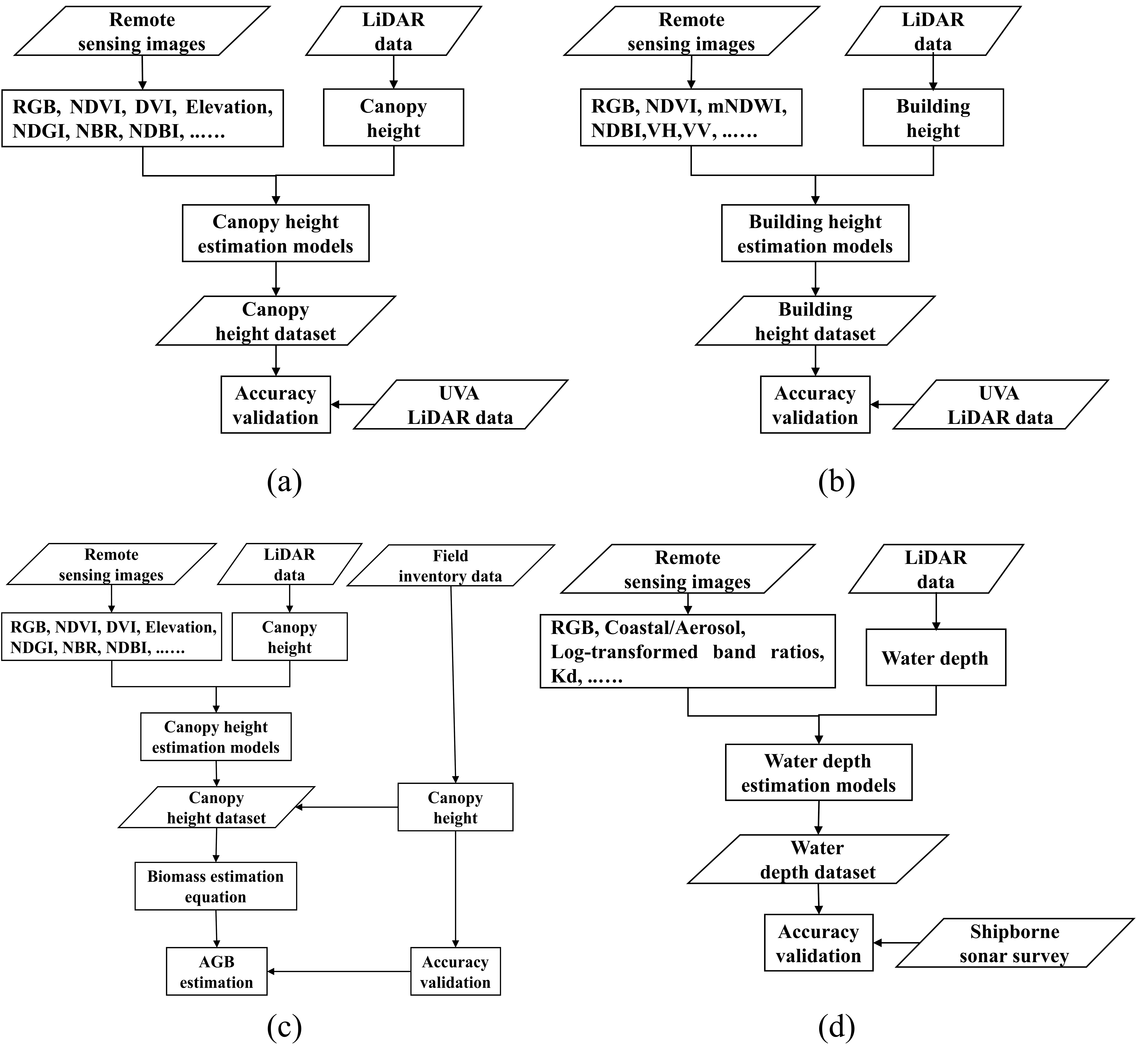}
    \caption{\textcolor{black}{Typical workflow of LiDAR-based inversion with weak supervision. (a) canopy height estimation, (b) building height estimation, (c) biomass estimation, and (d) water depth estimation.}}
    \label{fig:weaksignal}
\end{figure}

\begin{figure}[htbp]
    \centering
    \includegraphics[scale =0.45]{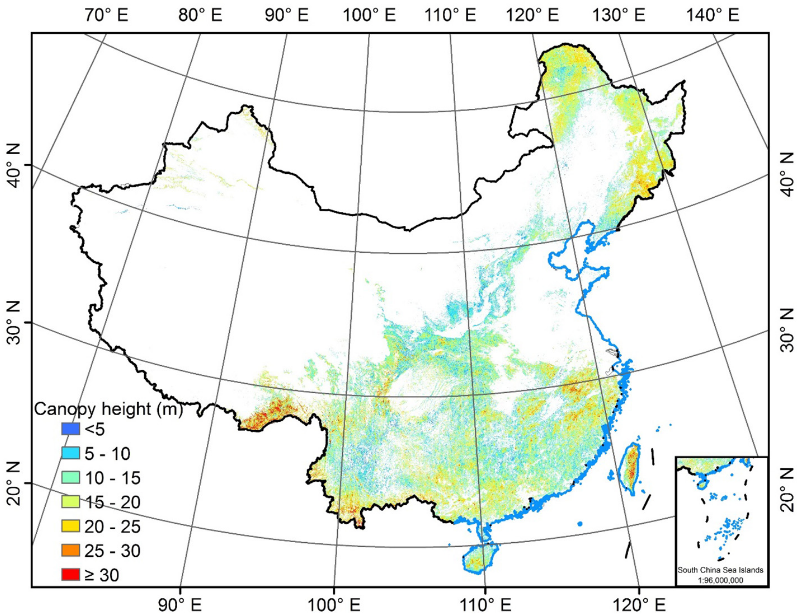}
    \caption{The NNGI-derived forest canopy height of China at 30 m resolution for 2019. The image is obtained from \cite{LIU2022112844}.}
    \label{fig:canopy height}
\end{figure}

\section{LiDAR Inversion with Weak Supervision}  \label{section_signals}

\textcolor{black}{In LiDAR remote sensing, LiDAR point clouds are often combined with remote sensing images for large-scale inversion tasks (e.g., canopy height, building height, aboveground biomass (AGB), water depth)}. Optical imagery provides continuous surface coverage with high revisit frequency and rich spectral information, while LiDAR data serves as supervisory signals, capturing structural and elevation details. However, due to the sparsity of LiDAR data, it cannot cover all inversion areas, making the generation of large-scale continuous maps challenging. Field surveys, which collect data such as tree diameter, height, by setting sampling points and survey regions, can provide more supervisory signals but are costly and resource-intensive. Additionally, regional differences and temporal variations can lead to domain shifts, which pose significant challenges to model accuracy and generalization. WSL is an effective approach to reduce costs and enhance model generalization. 
\textcolor{black}{In LiDAR inversion tasks, the application of WSL primarily follows two technical approaches. The first is domain generalization (DG), where a model is trained in a source domain rich with dense LiDAR data (e.g., from ALS coverage) and is then generalized to unseen target domains. The second is incomplete supervision, which directly uses spatially sparse LiDAR data (e.g., from GEDI and ICESat-2) as supervisory signals to train a model end-to-end over a large target area.
However, the application of incomplete supervision techniques remains limited
}
Table \ref{tbl:reversion_dataset} presents the commonly open datasets.

\textcolor{black}{In this section, we will review and summarize existing related work.
We categorize the applications in LiDAR remote sensing into specific areas, including canopy height mapping, building height estimation, biomass estimation, water depth estimation and other applications, as shown in Fig. \ref{fig:weaksignal}.}

\subsection{Canopy height estimation}

\begin{figure}[htbp]
    \centering
    \begin{subfigure}[b]{.41\textwidth}
        \centering
        \includegraphics[width=1.0\linewidth]{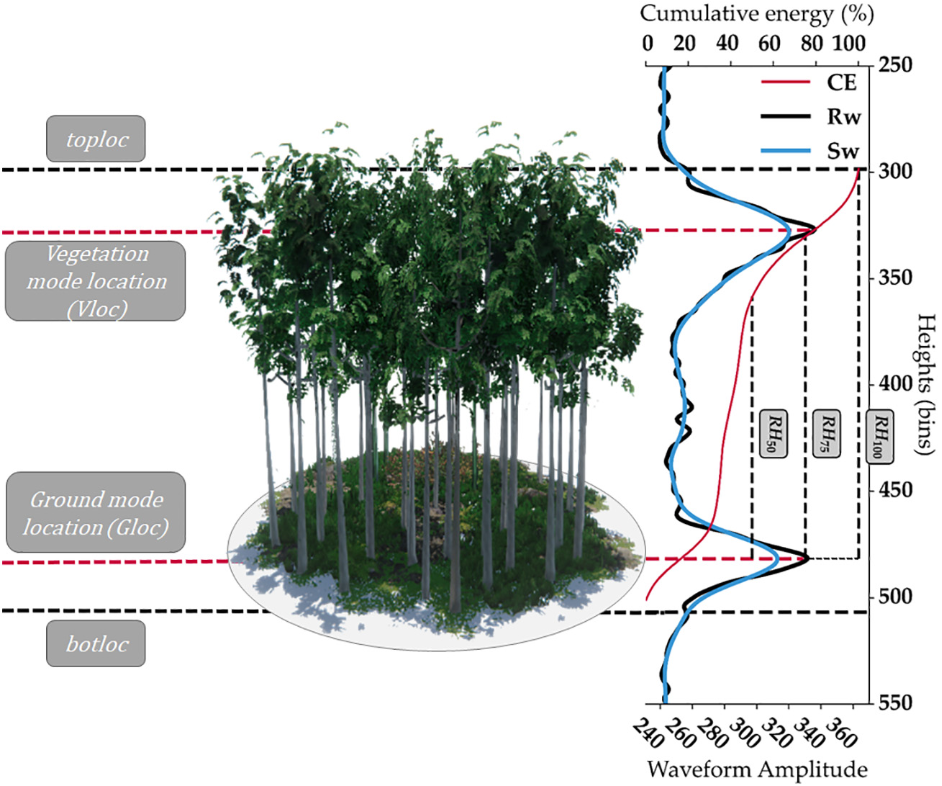}
    \end{subfigure}

    \begin{subfigure}[b]{.42\textwidth}
        \centering
        \includegraphics[width=1.0\linewidth]{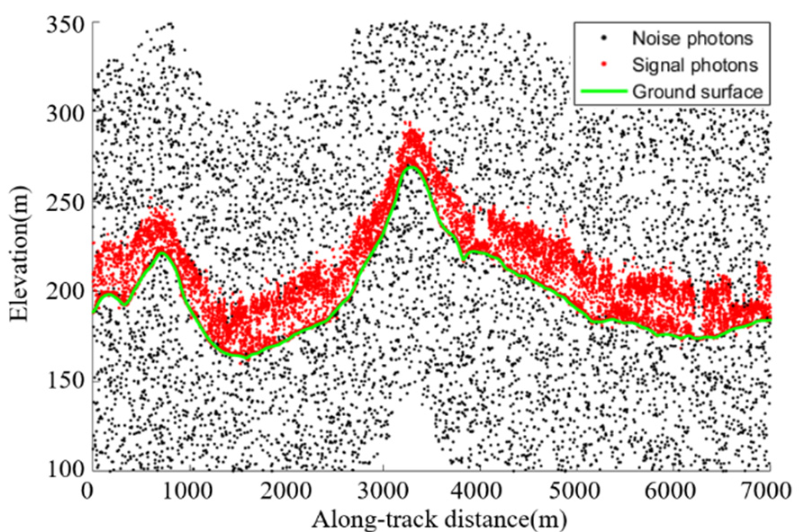}
    \end{subfigure}

    \caption{
 Example of an acquired GEDI waveform (Rw) over an Eucalyptus stand (Hdom = 25.9 m; V= 230.7 m3.ha-1), its smoothing (Sw) and corresponding waveform metrics (top). Strong beams of ICESat-2 in the daytime (bottom). These images are obtained from \cite{FAYAD2021112652,rs12203300}.}
    \label{fig:waveform}
 \end{figure}

Canopy height is a crucial parameter in forest ecosystems, playing a key role in assessing carbon storage, monitoring ecosystem health, and tracking land-use changes, as shown in Fig. \ref{fig:canopy height}. Accurate measurement of canopy height supports scientific research and informs forest management and conservation policies. Traditional ground-based measurements, such as forest inventories, offer high accuracy but are limited in coverage, time-consuming, and unsuitable for large-scale applications. Satellite remote sensing provides extensive coverage.
However, traditional multispectral and high-resolution imagery lacks the capability to capture 3D structural information like canopy height. 
LiDAR offers precise 3D structural data, but terrestrial and airborne LiDAR have limited spatial coverage.
Spaceborne LiDAR, such as GEDI and ICESat, offers global coverage, high data collection efficiency, low cost, and strong temporal consistency while providing high-precision 3D forest structural information \cite{zhu2022consistency}.
Fig. \ref{fig:waveform} illustrates how key ecological and terrain information is extracted from GEDI waveforms.

\begin{figure}[htbp]
    \centering
    \includegraphics[scale =0.12]{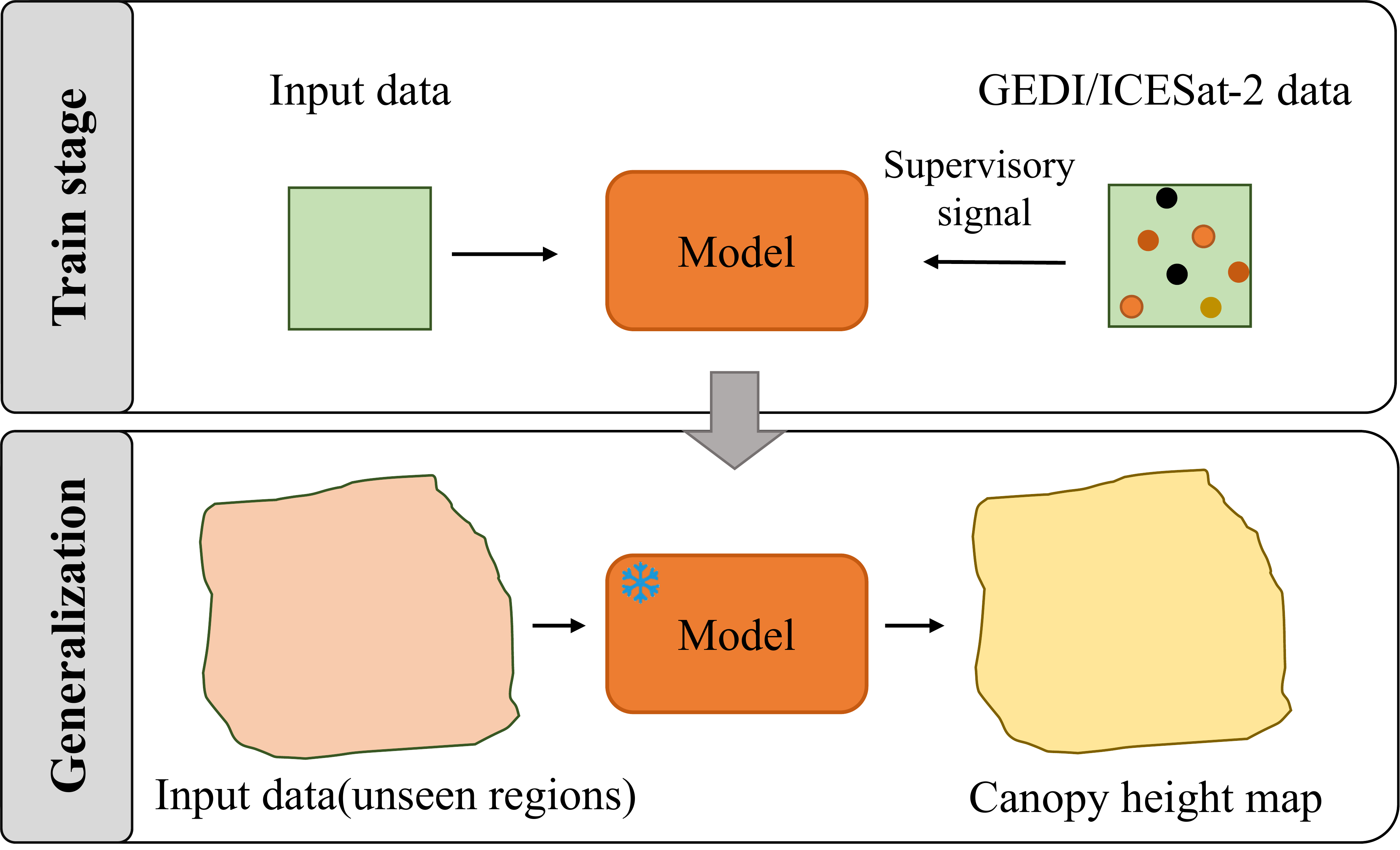}
    \caption{\textcolor{black}{Canopy height map generation based on remote sensing images and GEDI/ICESat-2 data. In the training phase, remote sensing images and corresponding GEDI/ICESat-2 data are used to train the neural network. In the extrapolation phase, the complete remote sensing data is input into the model to generate the full canopy height map.}}
    \label{fig:extrapolation}
\end{figure}

\begin{figure}[t]
    \centering
    \includegraphics[scale = 0.11]{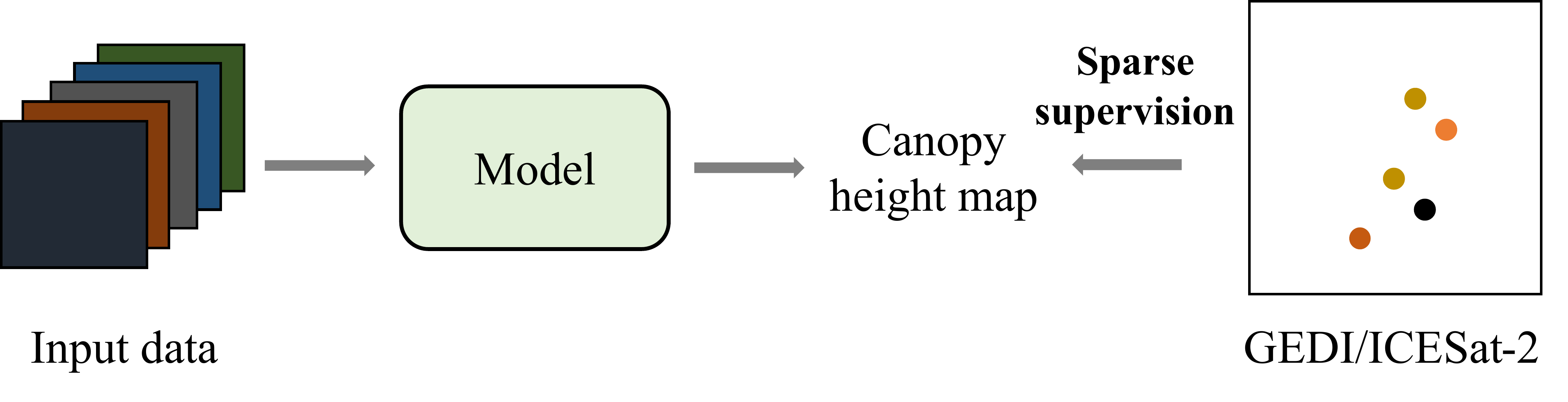}
    \caption{\textcolor{black}{Illustration of the model training process with sparse supervision from GEDI/ICESat-2 LiDAR.}}
    \label{fig:SpareGEDI}
\end{figure}


\textcolor{black}{Current approaches for canopy height retrieval typically use remote sensing imagery as training data and point clouds for supervision. One method follows the \textbf{domain generalization (DG)} setting, where a model trained on a small area is extrapolated to unseen regions, as shown in Fig. \ref{fig:extrapolation}.
This approach uses ALS data as labels, with GEDI/ICESat-2 LiDAR footprints and environmental auxiliary variables as inputs, to train a generalizable model for producing large-scale, high-precision canopy height maps. 
Another method follows the \textbf{incompletely supervised learning} setting, which uses sparse point clouds as weak supervision to train a model directly on large-scale remote sensing imagery, as shown in Fig. \ref{fig:SpareGEDI}.
This approach often employs a Convolutional Neural Network (CNN) that takes Sentinel-2 (S2) imagery and encoded geographic coordinates as input. Supervised by sparse GEDI LiDAR labels, the model concurrently estimates dense canopy height and its prediction uncertainty (variance).}

\textbf{Domain generalization}.
In the early stages, ALS point clouds are often used as weak supervisory signals.
For instance, 
Pourshamsi et al. \cite{POURSHAMSI202179} combine Polarimetric Synthetic Aperture Radar (PolSAR) parameters with limited  ALS data, using various machine learning algorithms to train on a subset of LiDAR samples and predict forest height over larger areas.
The emergence of spaceborne LiDAR has brought new opportunities.
\textcolor{black}{Some studies use ICESat-2 or GEDI data as supervisory signals and apply machine learning or deep learning to train on remote sensing images, ultimately mapping canopy height \cite{POTAPOV2021112165,LIU2022112844,CARCERERI2024114270}.}
For instance,
Potapov et al. \cite{POTAPOV2021112165} integrate NASA GEDI LiDAR data with Landsat multispectral time series data and use a machine learning regression tree model to train a forest canopy height prediction model within the limited coverage of GEDI.
They extend the model to a global scale, generating a 30-meter resolution global forest canopy height map for the year 2019.
Liu et al. \cite{LIU2022112844} integrate GEDI and ICESat-2 ATLAS data to map the spatially continuous distribution of forest canopy height in China. This method uses ALS data as training samples, trains the model on small regional areas, and then applies the model across the entire country to generate a 30-meter resolution forest canopy height map.
Wang et al. \cite{rs14153618} combine GEDI LiDAR data with other remote sensing data to improve the estimation of wall-to-wall canopy height. 
\textcolor{black}{
Turubanova et al. \cite{TURUBANOVA2023113797} utilize bagged regression tree models, integrating high-precision LiDAR data with long-term Landsat time-series imagery, to address the challenge of accurately mapping annual tree canopy extent and height dynamics across the entire European continent from 2001 to 2021.
Schwartz et al. \cite{schwartz2023forms} employ a U-Net model, using sparse but precise LiDAR height measurements from NASA’s GEDI mission as labels to train nationwide Sentinel-1 (radar) and Sentinel-2 (optical) imagery, and produce a 10-m-resolution forest canopy height map for France for 2020 that outperforms existing models in accuracy.
Similarly, they \cite{SCHWARTZ2024103711} also train a U-Net model that generates high-resolution forest canopy height maps of the whole``Landes de Gascogne'' forest.
Dixon et al. \cite{dixon2025canopy} employ a spatio-temporal deep learning model (ST-CNN) to analyze time series of high-resolution PlanetScope imagery, supplemented by radar and solar irradiance data, and produce 3-m resolution annual forest canopy height maps.
Zhang et al. \cite{zhang2025high} use an automated machine learning (AutoML) model to fuse multisource satellite imagery (Sentinel-1/2, Landsat 8, ALOS-2) with precise GEDI LiDAR measurements as labels to map high-accuracy, high-resolution tropical rainforest canopy height and aboveground biomass.
Xu et al. \cite{xiao202530} use precise but spatially discrete ICESat-2 LiDAR tree-height measurements as samples and train a Random Forest model to learn spectral and textural features from high-resolution GF-2 imagery, thereby producing citywide tree canopy height maps.
ARFCNet \cite{xiao2025ultra} combines convolutional layers, self-attention, and upsampling mechanisms to efficiently integrate airborne imagery, radar, and digital elevation models, enabling ultra-high-resolution (1 m) mapping of urban forest canopy height.
Su et al. \cite{su2025fused} train two independent deep learning models and fuse their predictions to generate high-resolution annual tree canopy height maps covering 20 years (2004–2024) for Italy.
Song et al. \cite{song2025seasonal} integrate LiDAR data with multi-source satellite remote sensing features to train a deep neural network model (STHNN) and employ the Shapley Additive Explanations (SHAP) method for feature selection, ultimately achieving large-scale mapping across the entire Shenzhen municipality.
Seppi et al. utilize \cite{seppi2025mapping} GEDI LiDAR data and SAOCOM radar imagery to train a machine learning model for generating high-precision forest canopy height maps.
Jamaluddin et al. \cite{11146436} propose a 3D spatio-temporal-spectral deep learning model that fuses Sentinel-1 time-series radar data with Sentinel-2 red-edge optical data to achieve high-precision estimation of canopy height over large-scale mangrove forests.
Malambo et al. \cite{malambo2024mapping} integrate high-precision laser altimetry data from the ICESat-2 satellite with Landsat and other multi-source auxiliary data, employing gradient boosting regression trees (XGBoost), to generate the first high-precision 30 m resolution vegetation canopy height map for the contiguous United States.
Wagner et al. \cite{wagner2024sub} employ a U-Net model to analyze high-resolution aerial imagery and reference LiDAR data, generating a high-precision, sub-meter resolution tree canopy height map across the entire state of California.
They also \cite{wagner2025high} train a U-Net model using high-resolution Planet satellite imagery and airborne LiDAR data to predict tree canopy heights across the entire Amazon rainforest.
Fogel et al. \cite{11095116} present Open-Canopy, a large-scale open benchmark dataset comprising SPOT satellite imagery from France and LiDAR height maps. Leveraging this dataset, they conduct comprehensive experiments and evaluations on several mainstream computer vision models—including U-Net and PVTv2—for the task of estimating forest canopy height from satellite imagery.
Pauls et al. \cite{pmlr-v235-pauls24a} achieve high-precision, high-resolution (10 m) global forest canopy height mapping by addressing key flaws in GEDI labels. They filter unreliable labels in steep terrain using SRTM data and train a U-Net on combined Sentinel imagery with a novel ``shift-resilient loss function'' to correct for systematic positional errors.
}

Deep learning techniques have made groundbreaking progress in recent years in 2D imagery, providing new solutions for the analysis of remote sensing data.
\textcolor{black}{Many studies employ deep learning frameworks under an \textbf{incompletely supervised learning} paradigm, treating point clouds as a weak supervisory signal for training on large-scale remote sensing imagery \cite{lang2023high,weber2024unifieddeeplearningmodel}.}
For instance,
Weber et al. \cite{weber2024unifieddeeplearningmodel} propose a unified deep learning-based model to predict global AGB density, canopy height, and canopy cover by fusing 13-channel inputs from Sentinel-1 and Sentinel-2 satellite imagery, multispectral data, and digital elevation models.
Fan et al. \cite{fan2024mappingcanopyheightprimeval} utilize a customized PRFXception neural network incorporating receptive fields at different scales, which learns canopy height information from sparse LiDAR sampling points and makes continuous inferences over the entire image coverage area, resulting in canopy height estimation at 10 m resolution.
Fayad et al. \cite{FAYAD2024113945} propose a hybrid visual transformer model called Hy-TeC to generate high-resolution tree canopy height maps using multi-source remote sensing data (Sentinel-1 and Sentinel-2) as inputs and GEDI LiDAR data as sparsely supervised ground truth. The model employs an architecture that combines a transformer encoder, a convolutional decoder, and a dual loss function of classification and regression to enhance the accuracy of the height intervals.
\textcolor{black}{Pauls et al. \cite{pauls2025capturing} train a 3D U-Net model to process satellite image time series accounting for seasonal variations, using sparse spaceborne LiDAR (GEDI) data as weak supervision labels, thereby generating high-resolution, annually varying dynamic tree canopy height maps across the European continent.
MARSNet \cite{chen2025multimodal} employs independent encoders tailored to sparse LiDAR, SAR, and optical imagery to extract modality-specific features, thereby enabling efficient fusion of these multi-source data to generate a continuous high-resolution forest canopy height map.
}

\textcolor{black}{The principal approach to canopy height estimation leverages sparse, pointwise height measurements from spaceborne LiDAR missions (primarily GEDI and ICESat-2) as weak supervisory signals. Early studies and many baseline comparisons relied on traditional machine-learning algorithms such as Random Forests. Contemporary approaches predominantly employ convolutional neural networks, such as U-Net and ResNet, to learn spatial context from image patches, while recent work increasingly adopts Transformer-based architectures, often in hybrid CNN and Transformer configurations, to capture long-range dependencies and improve accuracy in complex scenarios, for example, tall forest canopies. Despite substantial technical progress, the literature faces several common challenges: (1) data-quality issues, including label noise and signal saturation in input imagery; (2) limited model transferability, with models trained in one region generalizing poorly to other geographic contexts and forest types; and (3) persistent resolution mismatches that complicate multi-source data fusion.}

\begin{figure}[h]
    \centering
    \includegraphics[scale = 0.36]{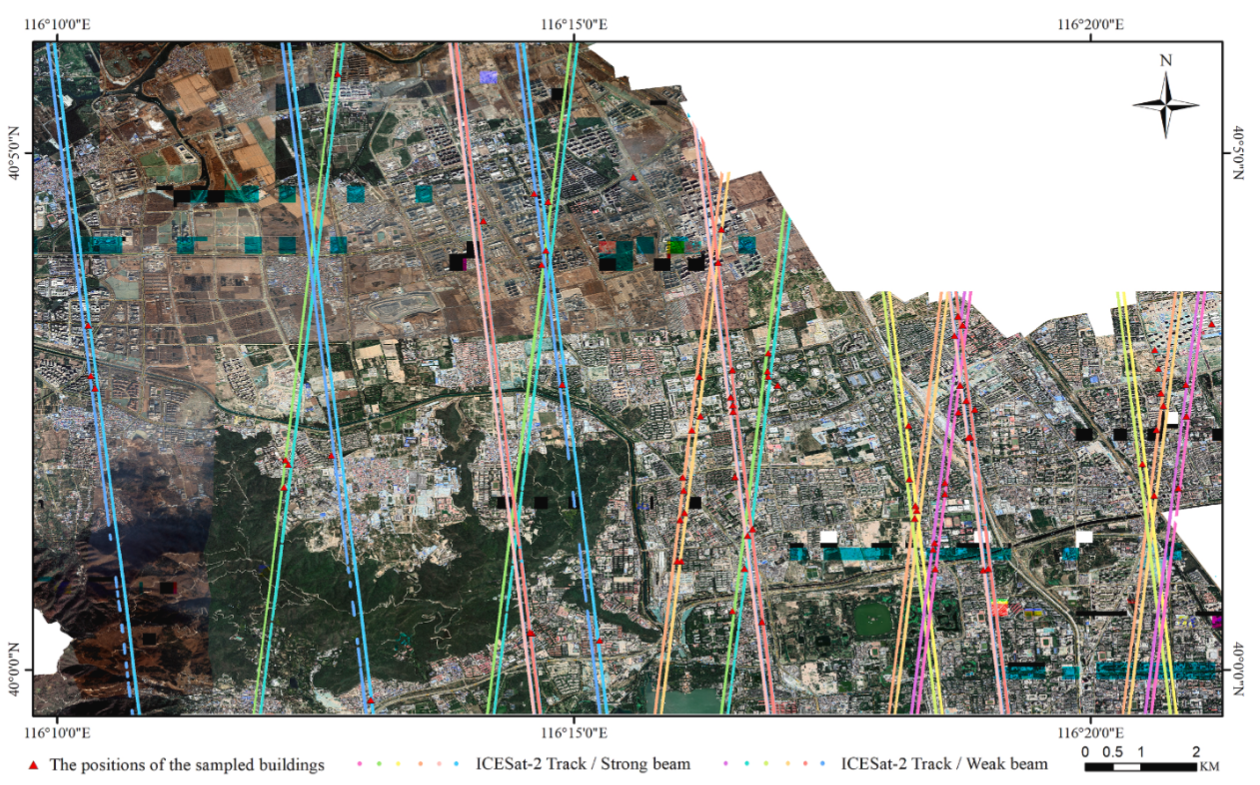}
    \caption{\textcolor{black}{Map of Haidian District, Beijing, with the ground tracks of ICESat-2 and the locations of field plots used to verify the accuracy of building height estimation. The 1-meter resolution satellite image is sourced from Google Earth and serves as the base map. The image is obtained from \cite{LAO2021102596}.}}
    \label{fig:BuildingHeight}
\end{figure}
\subsection{Building height estimation}

Building height, as a key parameter of urban 3D morphology, plays a critical role in energy consumption estimation, urban microclimate modeling, 3D reconstruction, population distribution analysis, and urban planning \cite{CAO2024114241}. Traditional methods, such as field surveys, are accurate but time-consuming and labor-intensive, making them unsuitable for large-scale applications. Remote sensing provides a cost-effective alternative but faces challenges: ALS data are expensive, SAR data struggle with single-building delineation, and shadow-based methods using high-resolution optical imagery, though practical, often rely on image metadata and building height annotations while ignoring azimuth variations, limiting their applicability. ICESat-2 satellite ATL03 photon data offer high-precision building height measurements but are restricted by orbital coverage, allowing measurement of only a limited number of buildings, as shown in Fig. \ref{fig:BuildingHeight}. 

Current methods integrate ICESat-2 photon data with remote sensing imagery to leverage their combined strengths, overcoming limitations of traditional methods and improving accuracy and applicability in building height extraction.
\textcolor{black}{Some studies adopt a \textbf{domain generalization (DG)} paradigm, training models on small-scale data and generalizing them to unseen large-scale areas \cite{LI2020111705,ma2023globalproductfinescaleurban}.}
For instance,
\textcolor{black}{
Zhou et al. \cite{zhou2022satellite} use precise in-situ building-height measurements from selected cities in North America and Europe as labels to train a physically-based regression model that interprets global Sentinel-1 SAR signals, thereby producing the world’s first high-resolution atlas of urban building heights.}
Ma et al. \cite{ma2023globalproductfinescaleurban} combine GEDI LiDAR data with features from Landsat-8, Sentinel-2, Sentinel-1, and terrain data to train Random Forest models across 68 subregions globally, generating a 150-meter resolution global urban building height map.
Ma et al. \cite{MA2023113392} combine GEDI LiDAR, optical (Landsat-8, Sentinel-2), and radar (Sentinel-1) data to produce a 150-meter resolution building height map for the Yangtze River Delta using a Random Forest model to scale discrete GEDI samples to regional mapping.
Kamath et al. \cite{Kamath2022GLOBUSGB} integrate LiDAR, ICESat-2, GEDI, and ALOS DSM data with machine learning to generate global building heights and urban canopy parameters. Using LiDAR-derived nDSM and features like building area and population density, they train a Random Forest model to predict building heights, which is then applied to cities worldwide for urban climate simulations, thermal comfort assessments, and energy consumption predictions.
\textcolor{black}{
Similarly, Tang et al. \cite{TANG2025114572} combine ICESat-2 elevation measurements with Sentinel satellite observations and a digital surface model (DSM), and employ a Random Forest regression to map built-up heights in the two megacities, New York and Shenzhen.
Ma et al. \cite{ma2024global} address the challenges of global urban building height data lacking timeliness and high resolution by employing a Random Forest model. They utilize precise spaceborne LiDAR (GEDI) data as training labels and integrate global multi-source satellite imagery (Landsat, Sentinel, etc.) as features to generate a new 150 m resolution global urban building height map for 2020.
}

\textcolor{black}{
Currently, large-scale building height estimation research has converged on a prevailing technical paradigm that leverages spaceborne LiDAR data as the principal weak supervisory signal for label generation, integrated with machine learning models and multi-source remote sensing imagery to enable spatially continuous mapping.
Specifically, training labels are primarily derived from two distinct spaceborne LiDAR systems. The first is GEDI, whose relatively large footprint (approximately 25 m) produces mixed relative height metrics that integrate information from the ground surface, vegetation, and built structures. The second is ICESat-2, whose denser photon data, when processed through advanced algorithms such as RANSAC, yield discrete yet more accurate estimates of building height. To support spatial extrapolation, models commonly incorporate time-series features derived from Sentinel-1 backscatter coefficients and Sentinel-2 optical imagery (for instance, shadow dynamics) as input variables. In terms of model selection, Random Forest remains predominant owing to its robustness in handling high-dimensional and heterogeneous data, while some studies employ Support Vector Regression (SVR) as an alternative.
}

\textcolor{black}{
Nevertheless, this paradigm faces intrinsic challenges related to the sparsity and quality of training samples. The linear sampling pattern of spaceborne LiDAR produces extremely sparse and spatially uneven observations, further affected by noise and geolocation errors that degrade label accuracy. This sparsity and imbalance, particularly the underrepresentation of tall buildings, often lead to a “saturation effect,” causing systematic underestimation of building heights exceeding approximately 30 m. Furthermore, models trained on sparse orbital tracks within a limited region struggle to capture diverse urban morphologies, thereby restricting spatial transferability and generalization. Finally, resolution mismatches in fusing multi-source and multi-scale remote sensing data (for example, 10 m imagery combined with 150 m grid data) constitute a major technical bottleneck for achieving high-precision building height estimation.}

\subsection{Biomass estimation}
\begin{figure}[t]
    \centering
    \includegraphics[scale = 0.32]{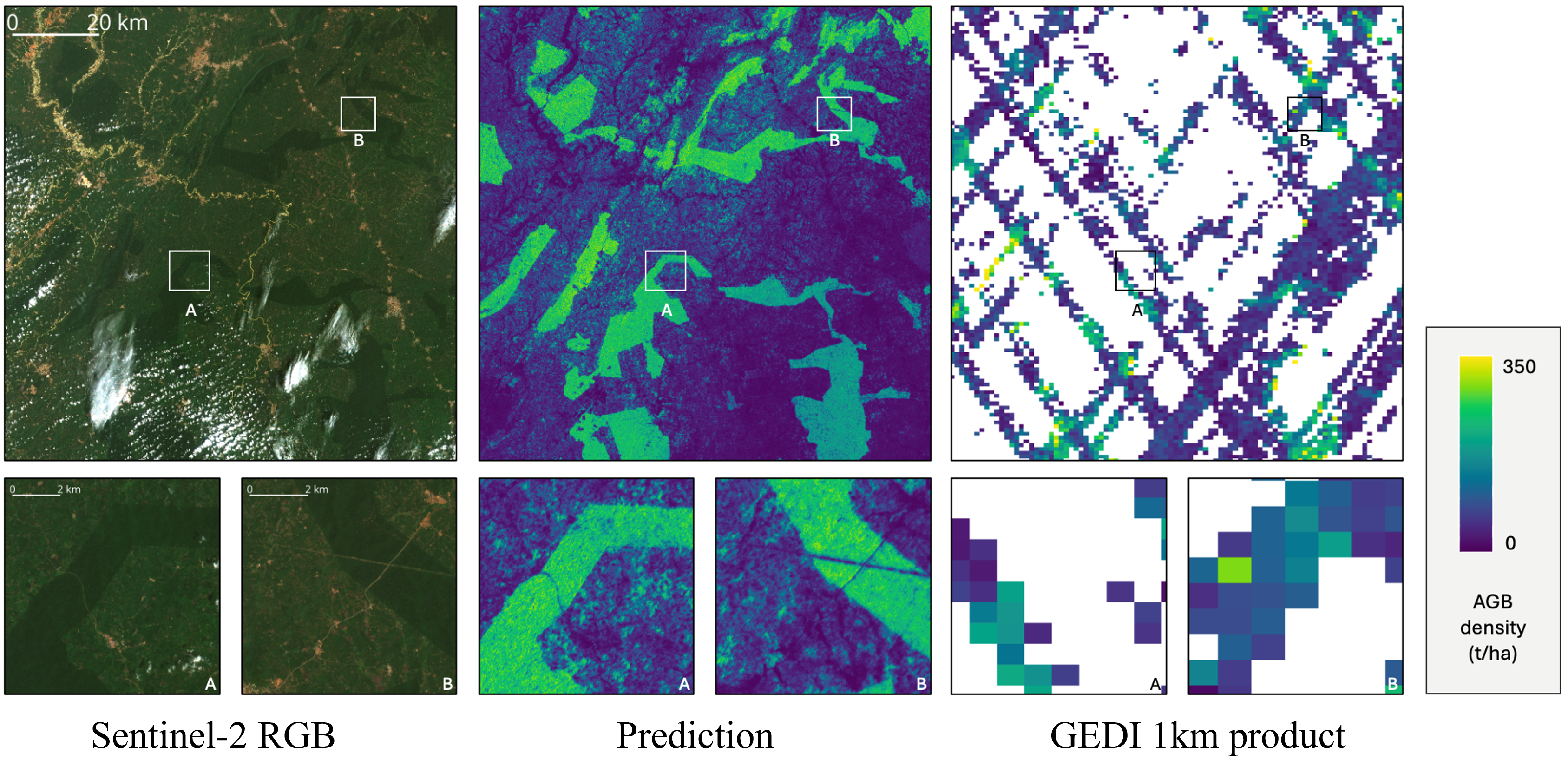}
    \caption{A high-resolution (10 m) map of AGB predictions. The image is obtained from \cite{sialelli2024agbdglobalscalebiomassdataset}.}
    \label{fig:AGB}
\end{figure}
Aboveground biomass (AGB) plays a crucial role in global climate change, particularly in carbon absorption and carbon sequestration \cite{sialelli2024agbdglobalscalebiomassdataset}. Accurate estimation of AGB is essential for sustainable forest management and biodiversity conservation, as shown in Fig. \ref{fig:AGB}. However, existing estimation methods, such as traditional destructive sampling and remote sensing, face challenges related to accuracy, coverage, and data scarcity. Spaceborne LiDAR provides high-precision biomass data, but its coverage is limited. By integrating multi-source Earth observation data and advanced machine learning methods, these challenges can be addressed to generate high-quality biomass maps.

\textcolor{black}{Current research typically employs regression models, machine learning, or deep learning techniques in conjunction with remote sensing imagery, airborne LiDAR (ALS), or satellite LiDAR data. Within a \textbf{domain generalization (DG)} paradigm, models are trained on small-scale regions and subsequently generalized to unseen spatial extents \cite{CAMPBELL2021112511,QIN2022106224,OEHMCKE2024113968}.}
Campbell et al. \cite{CAMPBELL2021112511} combine ground measurements with ALS data to construct a local-scale Random Forest model for predicting AGB within the ALS coverage. They then use the predictions of the local model as reference data and combine them with Landsat optical remote sensing data and GEDI LiDAR data to build a regional-scale model, enabling AGB estimation over a larger area.
Jiang et al. \cite{JIANG2022109365} combine ICESat-2 LiDAR data and Sentinel-2 optical remote sensing data and use an optimized Extreme Learning Machine (ELM) algorithm to estimate the AGB of natural forests in the eastern Tibetan Plateau. 
Vázquez-Alonso et al. \cite{rs14143432} combine ground surveys, LiDAR data, and Sentinel-2 imagery with a Random Forest model to classify forest types and estimate AGB. Using LiDAR-derived features, they build regression models to generate a high-resolution AGB map for a 25 km² area.
Duncanson et al. \cite{DUNCANSON2022112845} use ground and  ALS data from 74 global forest sites to generate training data with simulated GEDI waveforms. They develop stratified regression models based on relative height metrics, optimized by geographic and plant functional type (PFT) stratification, variable selection, and data transformation.
Deng et al. \cite{rs14225816} use ICESat-2 ATL08 LiDAR and multi-source remote sensing data (Landsat 8, Sentinel-1/2, DEM) with a Random Forest model to map 30-meter resolution canopy height. They further estimate forest AGB and carbon storage in Shangri-La City using height-DBH models and biomass formulas.
Oehmck et al. \cite{OEHMCKE2024113968} use deep learning models like PointNet \cite{qi2017pointnet}, KPConv \cite{thomas2019kpconv}, and Minkowski \cite{choy20194d} to estimate AGB and wood volume directly from  ALS point clouds. Leveraging 3D structural data, this method outperforms traditional statistical models and is applied to nationwide data in Denmark to create high-resolution biomass maps.
\textcolor{black}{
Nascetti et al. \cite{nascetti2023biomassters} train deep-learning models (e.g., U-Net and Transformer) using multimodal satellite time-series data (Sentinel-1 radar and Sentinel-2 optical imagery) as inputs and high-accuracy airborne LiDAR-derived aboveground biomass (AGB) as labels, thereby addressing the challenge of estimating forest biomass with high accuracy and fine spatial resolution.
Liu et al. \cite{doi:10.1126/sciadv.adh4097} combine high-resolution PlanetScope satellite imagery with airborne LiDAR data to train a U-Net deep-learning model and integrate allometric growth equations, thereby addressing the prior inability to accurately quantify carbon stocks of trees outside forests across Europe.
Guo et al. \cite{guo2023combining} fuse Sentinel satellite imagery to extend discrete GEDI LiDAR height points into a continuous forest canopy height map, which is further used to construct high-precision forest aboveground biomass (AGB).
Ma et al. \cite{ma2025forest}first integrate forest inventory data with ICESat-2 LiDAR data to generate high-precision aboveground biomass (AGB) sample points, then use these points as training labels to learn their relationships with wall-to-wall Landsat-8 imagery, topographic, and meteorological data via machine learning models, ultimately producing a continuous AGB distribution map across the entire study area.
Schwartz et al. \cite{schwartz2025retrieving} utilize a deep learning model, with GEDI data as labels, to train the model to directly learn from single-pass Sentinel-1 radar and Sentinel-2 optical satellite imagery and estimate forest height maps.
Bueno et al. \cite{bueno2025aboveground} utilize GEDI LiDAR biomass data as labels to train a machine learning model for generating a complete forest biomass map across the entire hurricane-impacted area.
Su et al. \cite{su2025canopy} first utilize U-Net to generate high-resolution canopy height maps, which are then input into a Random Forest model integrated with ground-measured data to ultimately estimate and map aboveground biomass.
Zurqani et al. \cite{zurqani2025multi} utilize the Google Earth Engine platform and select the gradient boosting tree (GBT) algorithm to fuse GEDI LiDAR, multi-source satellite imagery, and terrain data for estimating forest aboveground biomass.
Zhang et al. \cite{zhang2025correcting} fuse ICESat-2 LiDAR data to obtain optimal forest canopy height, which is incorporated as a key feature into a machine learning model, thereby significantly enhancing the accuracy of forest aboveground biomass estimation.
Cui et al. \cite{cui2025estimating} fuse GEDI with multi-source satellite remote sensing data (Sentinel-1 and Sentinel-2) using machine learning to provide more precise and detailed estimates of biomass consumption and carbon emissions for the 2020 Beachie Creek Fire than existing global products.
Contreras et al. \cite{contreras2025multi} successfully achieve precise estimation and mapping of aboveground biomass in large-scale olive orchards across the Mediterranean by integrating multi-source remote sensing data—including GEDI LiDAR, optical, and radar data—and employing machine learning models.
Cai et al. \cite{cai2025dynamics} use deep learning models integrated with multi-source remote sensing data to produce, for the first time, annual 30 m resolution maps of forest aboveground biomass across China from 1985 to 2023.
Considering that traditional biomass estimation models often overlook spatial autocorrelation (where nearby features are more similar) and heterogeneity (where relationships or patterns vary with spatial location) in geospatial contexts, Min et al. \cite{min2025improving} propose a novel machine learning model that integrates spatial correlation and heterogeneity. By combining sparse GEDI LiDAR data with continuous multi-source remote sensing imagery, they successfully map high-precision forest aboveground biomass across the western United States.
}

\textcolor{black}{In addition, a limited number of studies, under an \textbf{incompletely supervised learning} paradigm, employ satellite LiDAR point clouds as weak supervisory signals in combination with deep learning techniques for aboveground biomass (AGB) mapping.}
Weber et al. \cite{weber2024unifieddeeplearningmodel} develop a deep learning model to predict global AGB density, canopy height, and canopy cover using 13-channel inputs from Sentinel-1/2, multispectral data, and DEMs. With GEDI LiDAR data as labels, the model employs a multi-head CNN architecture to simultaneously predict these variables and their uncertainties, enabling high-resolution global forest monitoring.
Muszynski et al. \cite{muszynski2024finetuninggeospatialfoundationmodels} pre-train a Geospatial Foundation Model (GFM) with self-supervised learning, then fine-tune it on sparsely labeled data to estimate AGB in Brazil. Using a Swin-B Transformer as the encoder, the model combines labeled and unlabeled data to improve generalization, achieving AGB estimation accuracy comparable to U-Net while reducing computational costs.
\textcolor{black}{
Lyu et al. \cite{lyu2025forestcarbonnet} address the sparsity issue of GEDI LiDAR data by innovatively employing a point-supervised U-Net model to convert continuous multi-source remote sensing imagery into a high-resolution, seamless forest aboveground carbon density map.
}

\textcolor{black}{To enable large-scale, continuous mapping of forest aboveground biomass (AGB), recent studies primarily integrate multi-source remote sensing data with machine learning models. The core technical approach utilizes sparse footprint-level AGB density or canopy height data from NASA’s GEDI and ICESat-2 satellite LiDAR missions as weak supervisory signals, supplemented by diverse inputs such as Sentinel-2 or Landsat optical imagery, Sentinel-1 or ALOS-2 SAR backscatter, and SRTM-derived terrain variables, to achieve seamless mapping through machine learning. Among modeling approaches, Random Forest is most commonly employed due to its robustness in handling high-dimensional and heterogeneous data, while alternative models include Gradient Boosting Machines (GBM), Support Vector Regression (SVR), and convolutional neural networks (CNNs) such as U-Net, which more effectively capture spatial context. Despite significant progress, several key challenges remain, including signal saturation in high-biomass forests leading to underestimation, geolocation and spatial mismatches between sparse LiDAR footprints and high-resolution imagery, limited model transferability across ecological regions, and the complexity of multi-sensor data fusion caused by differences in spatial and temporal resolutions.}

\begin{figure}[htbp]
    \centering
    \includegraphics[scale =0.5]{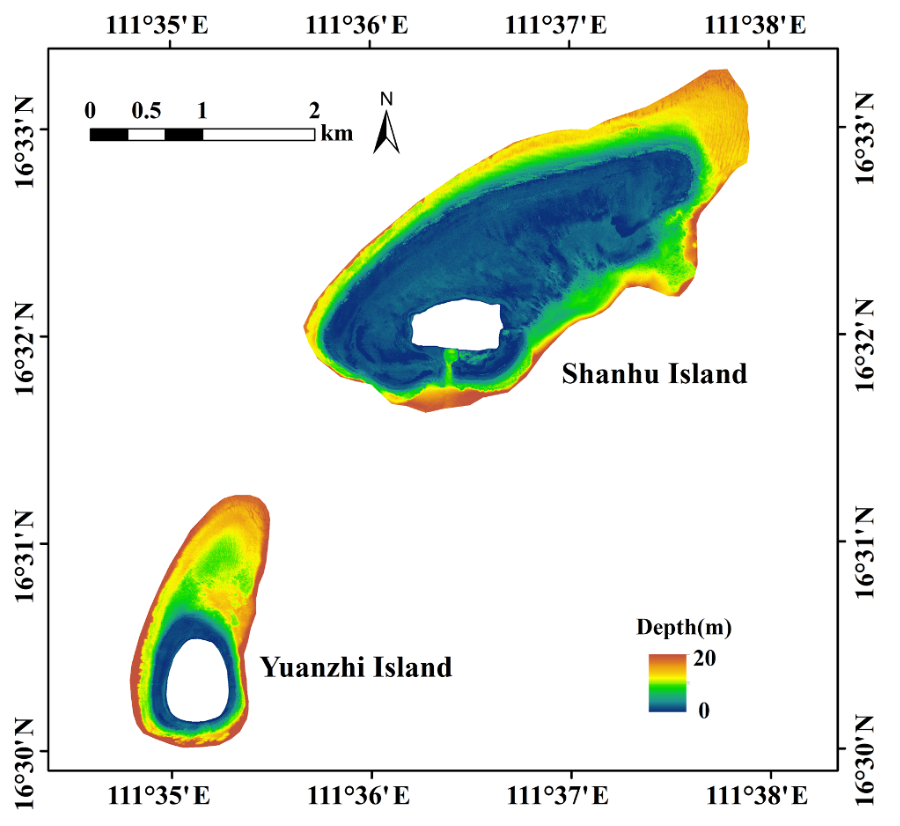}
    \caption{\textcolor{black}{Regional bathymetric map of Shanhu Island and Yuanzhi Island. The image is obtained from \cite{ZHAO2024104232}.}}
    \label{fig:water_depth}
\end{figure}
\subsection{Water depth estimation}
\textcolor{black}{
Traditional bathymetry relies on in-situ methods such as shipborne sonar, which are hindered by low efficiency and high labor and time costs, limiting their suitability for large-scale, high-frequency monitoring. Remote sensing inversion techniques provide a rapid, cost-effective alternative for obtaining underwater terrain data, thus garnering broad attention. In particular, ICESat-2's unprecedented high resolution and elevation accuracy enable revolutionary remote sensing bathymetry in the absence of field measurements, offering extensive application potential, as shown in Fig. \ref{fig:water_depth}.}

\textcolor{black}{
Existing studies predominantly employ \textbf{domain generalization} approaches. The dominant paradigm entails training machine learning or deep learning models using discrete LiDAR-derived labels together with continuous remote sensing imagery, and subsequently applying the trained models to large-scale bathymetric mapping.
Thomas et al. \cite{9834969} employ an extremely randomized tree regression model to learn the relationship between precise bathymetric points (labels) from ICESat-2 LiDAR and multispectral features from Landsat-8 imagery, thereby generating high-precision continuous bathymetric maps.
To enhance accuracy, Peng et al. \cite{9915464} propose a physics-assisted convolutional neural network (PACNN) that embeds water radiative transfer principles into the network architecture and uses precise ICESat-2 depth points as training labels for Sentinel-2 multispectral imagery, substantially improving the precision of traditional remote sensing bathymetric mapping.
Yang et al. \cite{YANG2023103310} innovatively integrate ICESat-2 along-track depth profiles with spectral trends from Sentinel-2 imagery to construct virtual ``free profiles'' that augment training data, then apply Random Forest and other machine learning inversion models to achieve high-precision lake-wide bathymetric mapping without in situ measurements.
Similarly, Othman et al. \cite{OTHMAN2025101432} combine Sentinel-2 multispectral imagery and ICESat-2 LiDAR data with AI methods such as quantile regression forests (QRF) to estimate shallow-water depths in Darbandikhan Lake, Iraq, without relying on conventional ground surveys.
For global lakes, Lv et al. \cite{rs17061052} leverage piecewise gradient boosting (piecewise GB) integrated with satellite imagery and LiDAR to deliver dynamic, continuous estimates of water depth and volume, particularly for small- to medium-sized lakes.
Gülher et al. \cite{gulher2025sensor} use multi-source inputs comprising optical satellite imagery, ICESat-2 LiDAR, and conventional echo-sounder data, along with Random Forest and XGBoost models, to overcome limitations in traditional shallow-water mapping and produce seabed bathymetry maps meeting IHO CATZOC A1 standards.
Du et al. \cite{DU2025102776} train a Kolmogorov–Arnold network (KAN) using precise ICESat-2 depth points and apply it to Sentinel-2 imagery to retrieve bathymetric maps comparable to those from in situ surveys.
Zhao et al. \cite{ZHAO2024104232} exploit the broad coverage of multispectral imagery and the sparse, high-precision measurements from spaceborne LiDAR, treating LiDAR data as exact labels to train an error-correction model that refines large-scale initial inversions and yields precise regional underwater topography.
Xie et al. \cite{rs16030511} develop a physics-informed convolutional neural network (PI-CNN) that incorporates radiative transfer principles, using ICESat-2 depth points as labels and wide-coverage Sentinel-2 multispectral imagery as input features to map nearshore water depths over extensive areas.
Zhang et al. \cite{11143589} utilize a Bayesian-optimized CatBoost model, training it with precise ICESat-2 depth data as labels to interpret Sentinel-2 imagery and address challenges in large-scale, automated, high-precision bathymetry for supraglacial lakes.
Liu et al. \cite{LIU2025104318} first perform seabed classification to decouple mixed signals from water and seabed, then apply a neural network model fusing ICESat-2 LiDAR data with Sentinel-2 optical imagery to tackle bathymetric inversion challenges in complex shallow-water environments.
}

\textcolor{black}{The integration of spaceborne LiDAR with optical imagery has become the prevailing paradigm for bathymetric inversion. This approach leverages high-precision, along-track sparse depth measurements from spaceborne LiDAR (e.g., ICESat-2) as reference labels together with spectral features derived from high spatiotemporal–resolution optical imagery (e.g., Sentinel-2) to train machine-learning models (e.g., Random Forest, convolutional neural networks) for regional, continuous depth estimation. Despite its considerable promise, the method faces several critical challenges in practice:
(1) Poor model transferability. Models calibrated for a given region are strongly dependent on the local inherent optical properties of the water and on substrate composition. When transferred to areas with different optical regimes, model performance commonly deteriorates, limiting generalizability.
(2) Spatial resolution mismatch among data sources. Scale discrepancies between the LiDAR laser footprint ($\approx$17 m for ICESat-2) and the pixel size of typical optical sensors (e.g., 10 m for Sentinel-2) introduce ambiguity when assigning sparse LiDAR depth labels to dense image pixels, thereby increasing label uncertainty.
(3) Data sparsity and physical limitations. The along-track sampling pattern of ICESat-2 produces incomplete spatial coverage. Furthermore, both active and passive optical signals attenuate rapidly in deep (>20–30 m) or highly turbid waters, constraining effective penetration depths and producing systematic underestimation of depths in deeper areas. Environmental factors such as sunglint and atmospheric contamination further degrade data quality.
(4) Difficulties in separating water-column and substrate contributions. Satellite-measured signals comprise a composite of water column backscattering and bottom reflectance. In environments with heterogeneous substrates (e.g., mixtures of sand, coral, and seagrass), reliably disentangling these components remains challenging and substantially affects inversion accuracy.
}

\subsection{Other applications}
\textcolor{black}{
\textbf{Canopy cover mapping.}
Canopy cover, defined as the proportion of ground obscured by the vertical projection of tree crowns, serves as a core indicator for assessing forest health, modulating ecosystem functions, and evaluating habitat quality.
Existing studies primarily employ generalization approaches, with the prevailing paradigm involving the training of machine learning or deep learning models using discrete LiDAR data as labels alongside continuous remote sensing imagery, followed by the application of these models to large-scale mapping.
Narine et al. \cite{NARINE2022113242} compare ICESat-2 satellite data against high-precision airborne LiDAR to assess two canopy cover estimation methods—direct computation and Random Forest—highlighting the superior performance of machine learning.
Viana-Soto et al. \cite{VIANASOTO2022102754} train a support vector regression (SVR) model with airborne LiDAR-derived labels for forest structure parameters (e.g., canopy cover and height), incorporating long-term Landsat imagery features to reconstruct historical metrics—including vegetation cover (VC), tree cover (TC), mean height (MH), and height variation coefficient (CVH)—spanning the past 30 years.
Similarly, Schlickmann et al. \cite{rs17020320} fuse multi-source data by using GEDI data and Sentinel-1 SAR and Harmonized Landsat-Sentinel-2 (HLS) optical imagery to train a Random Forest model, yielding the first high-resolution, seamless statewide canopy cover map for Florida.
}
	
\textcolor{black}{
\textbf{Carbon stock estimation.}
Forest carbon stock represents the total carbon absorbed and stored by forest ecosystems, playing a crucial role in mitigating global climate change. Traditional plot surveys calculate individual tree carbon stocks through field measurements of tree attributes combined with allometric equations, offering high accuracy but incurring substantial costs and scalability challenges. Remote sensing methods, by contrast, leverage satellite imagery—such as optical or radar data—to construct models that estimate regional-scale carbon stock distributions, affording extensive spatial coverage.
Existing studies primarily employ generalization approaches, with the prevailing paradigm involving the training of machine learning or deep learning models using discrete LiDAR data as labels alongside continuous remote sensing imagery, followed by the application of these models to large-scale mapping.
Jiao et al. \cite{rs15051410} fuse Sentinel-2 optical imagery with sparse GEDI data using a fully convolutional network (FCN) to generate high-resolution continuous forest canopy height maps. They then apply an airborne LiDAR (ALS)-derived canopy height–carbon stock model to convert these maps into high-precision carbon stock distributions for the project area, from which they estimate forest carbon emission reductions.
Sun et al. \cite{rs16203847} integrate limited field plots with high-precision ALS data through a Bayesian hierarchical model to produce reliable carbon stock samples in representative areas; these serve as high-quality labels to train a Random Forest model, facilitating high-precision mapping of forest carbon stocks across larger regions using broad-coverage Sentinel-2 and auxiliary data.
}

\textcolor{black}{
In addition,
Araza et al. \cite{araza2023spatial} employ an ensemble machine learning model (integrating Random Forest, gradient boosting trees, and support vector machines) to learn relationships between multi-source satellite remote sensing data (e.g., global biomass and forest height change maps) and precise carbon flux labels from national forest inventories and airborne LiDAR, thereby generating continuous \textbf{carbon flux maps} for five countries spanning 2010–2018 to support United Nations carbon accounting.
Xi et al. \cite{Xi31122022} leverage phenological differences between overstory evergreen conifers and understory deciduous vegetation, training a support vector regression model with multi-temporal Sentinel-2 optical imagery and GEDI LiDAR data to enable large-scale mapping of spatially continuous \textbf{understory vegetation density}.
Similarly, Yu et al. \cite{Yu03072023} propose a large-scale method for estimating \textbf{forest growing stock volume}, using high-precision airborne LiDAR (ALS) data as a ``bridge'' to generate reliable training labels for Sentinel-2 imagery and producing high-accuracy maps of standing timber volume for larch forests across China via a Random Forest model.
Yao et al. \cite{YAO2024104010} train a Random Forest model to capture complex relationships between ICESat-2 satellite elevation points and surface features from Sentinel-1/Sentinel-2 imagery, generating continuous high-precision \textbf{terrain elevation maps} at 10 m resolution.
}

\textcolor{black}{
These studies typically employ a domain generalization strategy, where models are first trained on small-scale regions and subsequently generalized to unseen large-scale datasets. They leverage spatially sparse yet high-precision, weakly supervised signals from spaceborne LiDAR (e.g., GEDI, ICESat-2) footprints to provide accurate measurements of vertical forest structure, such as canopy height and elevation, despite their discontinuous coverage.
Concurrently, ALS data often serve as a bridge to upscale local ground-truth plot data, generating denser pseudo-labels for model training.
In terms of model selection, Random Forest is widely adopted due to its effectiveness in handling high-dimensional data, its robustness against overfitting, and its stable performance.
Overall, the primary research challenges include the signal saturation of optical remote sensing (which makes it difficult to distinguish subtle biomass differences in dense vegetation), the spatiotemporal mismatch of multi-source data (such as alignment errors between LiDAR footprints and satellite pixels), and the insufficient representativeness of reference data.
These factors introduce significant uncertainty when extrapolating results to unsampled regions, highlighting fundamental challenges in domain adaptation and generalization for large-scale LiDAR–remote sensing.
}

\subsection{Summary and discussion}

\textcolor{black}{LiDAR-based inversion seeks to generalize from sparse spatiotemporal observations to continuous, large-scale parameter maps—a problem that is essentially domain generalization (DG): models are learned from limited measurements in observed regions or periods and applied to unseen domains. supervisory signals typically come from two hierarchical sources: (i) \emph{in situ} ground measurements (e.g., tree height, biomass, canopy cover), which provide the most accurate references; and (ii) LiDAR data (e.g., GEDI, ALS), which supplies structurally informative signals bridging sparse measurements and continuous optical imagery.
These complementary properties make LiDAR an effective weak supervision source for large-scale inversion of biomass, tree height, water depth, canopy cover, and carbon stock.}

\textcolor{black}{However, current inversion pipelines face coupled data–model limitations. On the data side, LiDAR samples are spatially incomplete and unevenly distributed, subject to geolocation errors and canopy-structure–dependent biases; RH metrics are statistical proxies rather than physical ground truth, and ICESat-2 exhibits increased sensitivity to terrain and noise. A frequently overlooked issue is \emph{resolution mismatch}: for instance, a single GEDI footprint (~25 m diameter) is often assigned to the center pixel of a 10 m Sentinel-2 image (or finer), implicitly assuming within-footprint homogeneity and ignoring sub-footprint heterogeneity. Similar mismatches between airborne LiDAR point clouds and imagery introduce systematic supervision noise and propagate uncertainty through training.}

\textcolor{black}{On the modeling/method side, many large-scale studies still rely on simple statistical or classical ML models (e.g., linear regression, Random Forest, Gradient Boosting) that are easy to train under sparse labels but offer limited cross-domain generalization. Although CNNs and CNN–Transformer hybrids are increasingly used to exploit spatial context (e.g., Landsat/Sentinel plus DEM), label uncertainty and domain shift are often not treated explicitly. In practice, model transferability is typically enhanced through heuristic and experience-based strategies: incorporating empirical biogeophysical variables (e.g., NDVI, topography, tidal dynamics), region-specific training or local fine-tuning, temporal stratification to reduce seasonal domain shifts, and multi-source data fusion (e.g., combining ICESat-2, optical imagery, SAR, and tidal information). Temporal sample migration is also used to enlarge training sets and improve robustness across environmental conditions. While these strategies can enhance performance within local regions, they depend heavily on expert knowledge, manual rules, and task-specific tuning, which limits scalability and reproducibility. As a result, most existing applications achieve only \emph{regional or site-specific} generalization, and their accuracy tends to degrade substantially when extended to broader spatial or temporal domains.}

\textcolor{black}{In contrast, advances in machine learning offer systematic, learnable DG solutions—such as adversarial domain alignment, domain-invariant feature learning, test-time adaptation, meta-learning, and uncertainty-aware representation learning—that can replace or augment these heuristic strategies. Weak supervision provides the algorithmic foundation for treating LiDAR observations as imperfect signals rather than perfect labels: pseudo-label generation, consistency regularization, and contrastive learning improve robustness to sparse and noisy supervision; multi-modal fusion and self-training support scale transfer; active learning can prioritize informative LiDAR samples. Combining weak supervision with DG/DA/TTA enables models to adapt to spatiotemporal heterogeneity, resolution mismatch, and label uncertainty in a principled and scalable manner. This shift from heuristic rules to learnable generalization represents a critical next step toward robust, transferable, and cost-efficient LiDAR-based inversion.}

\begin{figure}[t]
    \centering
    \includegraphics[scale = 0.15]{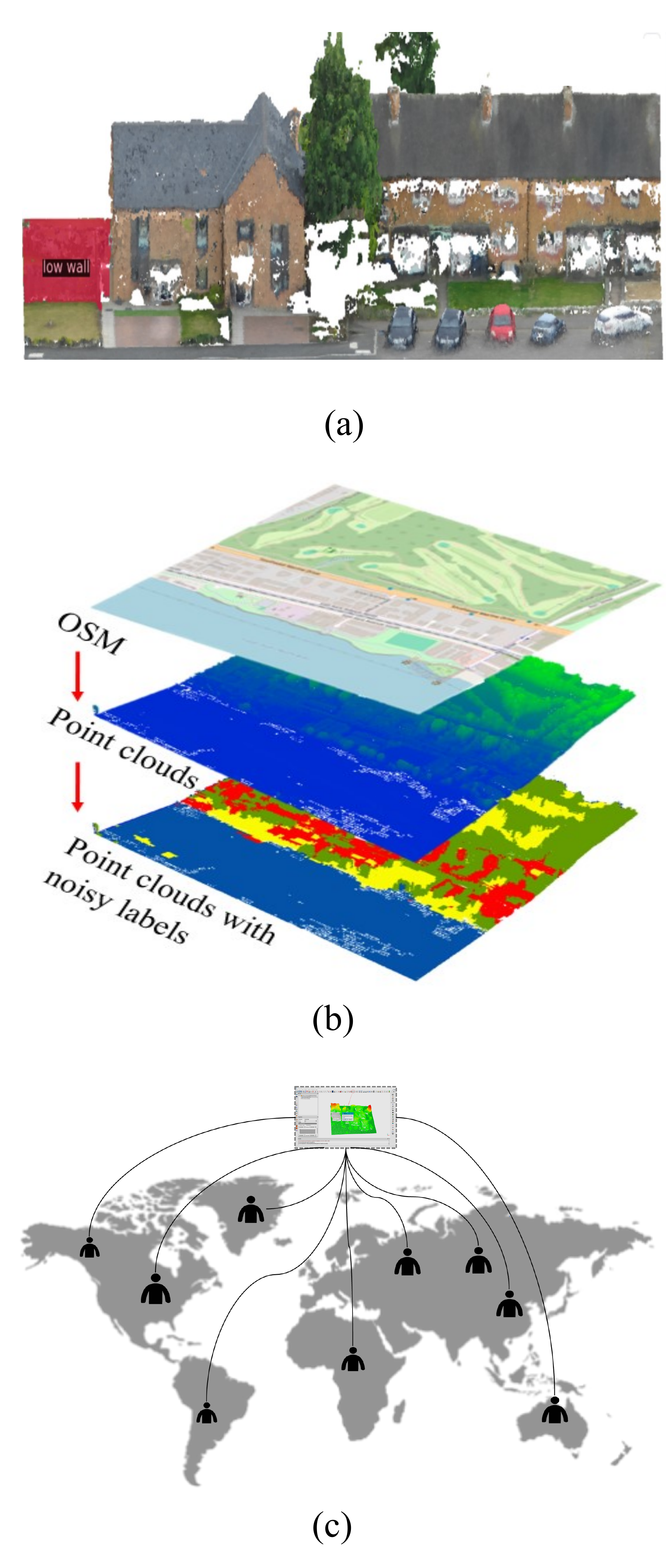}
    \caption{Different annotation methods. (a) based on foundation models \cite{NEURIPS2023_3ef61f7e}, (b) based on historical maps, and (c) based on crowd-sourced data. }
    \label{fig:generalannotation}
\end{figure}
\section{Perspectives and Future Work}  \label{section_future}

\subsection{\textcolor{black}{Challenges of applying WSL to LiDAR remote sensing}}

\textcolor{black}{
\textbf{Uniqueness of LiDAR remote sensing data.} Compared to natural images and remote sensing imagery, LiDAR remote sensing differs fundamentally not only in data organization (raster grids vs. point clouds/waveforms) but also in exhibiting more complex and challenging characteristics. First, point cloud sampling is influenced by multiple factors such as altitude, scanning angle, echo energy, and terrain occlusion, resulting in highly uneven density with sparse, irregular, and spatially discontinuous features; this contrasts sharply with the fixed resolution of imagery and the high density and local stability of indoor point clouds. Second, the wide range of observation platforms (e.g., drones) leads to substantial variations in viewpoint and scale, causing much greater heterogeneity in observation geometry and object structures than in imagery or indoor scenes. Third, annotation is significantly more difficult due to diverse object types, complex terrain undulations, and occlusions, whereas indoor point clouds benefit from simple structures and controllable scenes, facilitating standardized and automated labeling. Finally, LiDAR domain variations are strongly coupled with geographic environments, seasons, and ecological dynamics, with temporal changes markedly amplified; in contrast, domain differences in natural images primarily stem from illumination styles, remote sensing imagery is governed by imaging geometry, and indoor point clouds exhibit the most stable environments.}

\textcolor{black}{
In summary, the challenges of LiDAR remote sensing in sampling sparsity, geometric scale variations, annotation complexity, and spatiotemporal domain heterogeneity far exceed those of images and indoor point clouds, directly increasing the difficulty of algorithm modeling.}

\textcolor{black}{\textbf{Limitations of existing LiDAR WSL.} These characteristics dictate that weakly supervised learning for LiDAR remote sensing cannot simply adopt techniques from images or indoor point clouds. First, the discrete and sparse nature of point clouds disrupts the spatial continuity assumptions underlying image-based weak supervision, making label propagation and pseudo-label generation unstable in discontinuous regions, whereas indoor point clouds are less affected due to their density and controllability. Second, LiDAR lacks spectral and textural priors, with dominant information being geometric structures and intensity, rendering image-based weak supervision methods reliant on saliency or attention difficult to transfer; indoor point clouds often incorporate RGB data, enabling direct leverage of visual foundation models. Third, its domain variations and spatiotemporal dynamics are far stronger than in images and indoor point clouds, imposing higher demands on model robustness and generalization. LiDAR has a more urgent need for domain generalization (DG), yet research in this direction remains scarce; in contrast, the image domain has mature methods and pre-trained models (e.g., CLIP, SAM) to mitigate domain shifts, while LiDAR lacks comparable support. Finally, LiDAR's unique waveform-point cloud multimodal features (e.g., waveform decomposition, multiple echoes, sub-footprint) further increase uncertainty in data interpretation and pseudo-label modeling.}

\textcolor{black}{
Therefore, the weak supervision problem in LiDAR remote sensing is not merely ``insufficient labeling'' but a systemic challenge driven by sampling structures, feature modalities, spatiotemporal dynamics, and task mechanisms, necessitating the development of a dedicated weak supervision theoretical and methodological framework independent of images and indoor point clouds.}

\subsection{\textcolor{black}{LiDAR remote sensing meets foundation models}}

\textcolor{black}{Foundation models, characterized by large-scale, multi-source pre-training, are driving a paradigm shift in artificial intelligence. These models, including Large Language Models (LLMs) and Vision-Language Models (VLMs), demonstrate remarkable generalization and zero/few-shot learning capabilities, enabling adaptation to a wide range of downstream tasks through fine-tuning. LLMs excel at the deep understanding and generation of natural language \cite{zhao2023survey}, while VLMs establish profound connections between visual content (images, videos) and textual descriptions by aligning visual and textual modalities.
}

\textcolor{black}{In the field of Earth observation, the deep integration of LiDAR technology with foundation models is forging a new path for 3D perception. This convergence combines the precise 3D geometric measurement capabilities of LiDAR with the powerful deep semantic reasoning of foundation models, advancing 3D scene understanding from traditional classification of predefined categories toward an open-world cognitive intelligence paradigm that supports natural language interaction. This fusion exhibits significant advantages in two core areas:
Parameter Inversion: In tasks such as forest biomass estimation, high-precision geometric priors from LiDAR (e.g., canopy structure) effectively constrain numerical models. When combined with the semantic context and domain knowledge extracted by VLMs/LLMs, the accuracy, robustness, and generalization of model estimations across different regions and sensors are significantly enhanced \cite{xue2024reovlmtransformingvlmmeet}.
Scene Interpretation: For tasks like semantic segmentation and object detection, the open-vocabulary characteristic of VLMs overcomes the reliance of traditional models on fixed label sets. This enables open-set recognition and rare object detection, dramatically increasing the flexibility and intelligence of scene interpretation \cite{cen2022open}.
}

\textcolor{black}{
However, transferring the powerful semantic capabilities of foundation models to sparse, irregular LiDAR 3D point cloud data remains a formidable challenge. Mainstream foundation models are predominantly trained on 2D regular image grids and 1D text sequences, rendering their network architectures incompatible with direct point cloud processing. This incongruity leads to two primary issues:
Modality Mismatch and Projection Error: The significant discrepancy between the 2D image-based pre-training of foundation models and the sparse 3D geometry of LiDAR point clouds results in performance degradation. Furthermore, projecting point clouds onto 2D planes (e.g., elevation maps, bird's-eye views) inevitably introduces geometric information loss, occlusion distortion, and resolution limitations, which further compromises the accuracy of downstream tasks.
Cross-Modal Alignment Uncertainty and Domain Shift: When transferring knowledge using registered image-LiDAR data pairs \cite{zhang2024opensight}, noise and semantic ambiguity from weak labels, coupled with unreliable alignment due to temporal, viewpoint, resolution, and geometric registration discrepancies, diminish the effectiveness of knowledge transfer. Moreover, significant domain shifts, arising from sensor heterogeneity and geographic variability, severely constrain the generalization performance of these models.
}

\textcolor{black}{To systematically bridge the modal gap between LiDAR and foundation models, the academic community is actively exploring three emerging strategies:
(i) Prompt/Adapter-based Fine-tuning: This approach involves lightweight adjustments to inject structural cues extracted from LiDAR (e.g., canopy height profiles) as geometric prompts into pre-trained models.
(ii) Cross-modal Pre-training: This strategy aligns the geometric features of LiDAR with the semantic features of imagery within a shared representation space during the pre-training phase, using methods like contrastive learning or knowledge distillation.
(iii) Pseudo-label Refinement: This iterative process fuses coarse semantic labels generated by VLMs with precise geometric information from LiDAR to produce high-quality supervisory signals that, in turn, refine the model's training.}

\textcolor{black}{
Within this new paradigm, the role of weak supervision extends beyond merely addressing label scarcity; it is elevated to a critical bridge connecting LiDAR's geometric information with the semantic space of foundation models. Weak supervisory signals derived from LiDAR (e.g., canopy height, digital elevation models) serve as a low-cost source of rich structural information. They not only anchor the cross-modal feature alignment but also inject essential geometric priors into the high-dimensional semantic space. This guides large models to comprehend 3D geometric structures without requiring massive volumes of pre-trained point cloud data. For instance, AlphaEarth Foundations utilizes sparse GEDI LiDAR data (for forest canopy height) as a reconstruction target within its self-supervised learning framework. This process enables the model to leverage other, denser data sources, such as optical and radar imagery, to learn and generate continuous, high-resolution, global estimates of vegetation height \cite{brown2025alphaearthfoundationsembeddingfield}. In return, foundation models endow LiDAR analysis tasks with unprecedented semantic breadth and open-world reasoning capabilities. This synergistic cycle of geometric priors and semantic capabilities establishes a bidirectional pathway for knowledge transfer, laying a solid foundation for the ultimate development of point cloud-native foundation models.}

\subsection{\textcolor{black}{Future directions for LiDAR remote sensing}}

We have identified several promising research directions in the domain of weak supervision for LiDAR remote sensing:  

\textbf{Leveraging Ubiquitous Annotations.} 
The development of large-scale models, including generative models and vision-language foundation models, has enabled the generation of vast quantities of annotations \cite{dong2023leveraginglargescalepretrainedvision}. These models have already demonstrated significant success in annotating remote sensing imagery. Beyond these automated approaches, numerous sources of annotations are readily available, including crowd-sourced data \cite{Russakovsky2014ImageNetLS}, network-generated labels \cite{9879317}, historical maps (e.g., OpenStreetMap \cite{9119753}), and annotations derived from remote sensing images, as shown in Fig. \ref{fig:generalannotation}.
This abundance of annotation sources represents a valuable opportunity for advancing weakly supervised LiDAR data interpretation. Future research should focus on fully harnessing these ubiquitous annotations to enhance the precision and efficiency of weakly supervised LiDAR-remote sensing tasks. By integrating these diverse annotation sources, researchers can reduce reliance on labor-intensive manual labeling, improve model performance, and significantly lower operational costs. The challenge lies in developing robust frameworks that can effectively utilize such diverse and sometimes noisy annotations to achieve consistent and scalable outcomes.

\textbf{Domain Variability in LiDAR Remote Sensing.} 
The dynamic nature of these phenomena—driven by temporal, seasonal, and regional variability—exacerbates the challenges of domain shifts \cite{Wang2024TesttimeAF}. These challenges are further amplified by the inherent modality inconsistencies among LiDAR systems, such as ALS, TLS, and spaceborne sensors, which differ in resolution, coverage, and acquisition geometry. The combination of domain variability and multimodal inconsistencies makes obtaining comprehensive, consistent annotations both expensive and infeasible, particularly given the high cost and complexity of point cloud labeling.

WSL offers a practical solution by leveraging limited labeled data and vast unlabeled datasets to mitigate annotation scarcity. Techniques such as semi-supervised learning \cite{7325695,10660507}, feature alignment \cite{Luo2020UnsupervisedSA}, and pseudo-labeling \cite{9915611} are particularly suited to LiDAR data, utilizing its geometric and spatial properties. When combined with DA, these methods can address not only temporal and spatial domain shifts but also the modality gaps inherent to LiDAR systems.

Future LiDAR research must prioritize annotation-efficient and modality-robust approaches to handle domain and sensor variability. Weak supervision has the potential to enhance model generalization and adaptability, enabling LiDAR to address dynamic Earth observation challenges such as disaster response, urban monitoring, and environmental change analysis. By bridging technical gaps, WSL can ensure LiDAR's broader applicability in real-world, rapidly evolving scenarios.

\textbf{LiDAR Remote Sensing Meets Open World.} 
Current LiDAR remote sensing technologies have made significant progress in LiDAR data interpretation and LiDAR-based inversion. However, most existing methods focus on closed-set tasks, where neural networks are trained on a specific dataset and tested on the same category set. While effective for recognizing known classes, this approach struggles with the dynamic and uncertain nature of the real world, including unseen categories, environmental changes, and new tasks \cite{li2024segearthovtrainingfreeopenvocabularysegmentation}. As a result, traditional closed-set methods are limited in addressing the diversity and complexity of the open world. Existing LiDAR datasets are relatively small and fixed in terms of categories. Future research should focus on developing LiDAR datasets with richer, more diverse categories that better reflect real-world scenarios. Additionally, integrating foundational models and multimodal data to develop cross-modal, generalizable, and scalable models will be a key direction for future work.

\textbf{New Paradigms for LiDAR Remote Sensing Inversion.} 
Most current approaches in LiDAR remote sensing involve extracting sparse signal-derived features, such as elevation or vegetation metrics, and using them to construct regression or classification models for large-scale remote sensing retrieval tasks. However, with the advancement of deep learning, there is growing interest in leveraging weak supervision to fuse LiDAR data with optical remote sensing imagery for improved retrieval accuracy \cite{lang2023high}.  Constructing spatiotemporally continuous remote sensing retrieval products using sparse supervisory signals (e.g., LiDAR) has emerged as a promising research direction \cite{Li2024LearningSD}. This emerging paradigm raises two critical challenges: 1) how to effectively integrate multi-modal remote sensing data for retrieval tasks, and 2) how to utilize sparse LiDAR data as a weak supervisory signal to train deep learning models for large-scale retrieval.  

Meanwhile, the use of LiDAR as a weak supervisory signal introduces unique challenges in training deep learning models. Sparse LiDAR data, despite its limitations, provides valuable ground truth for guiding retrieval tasks across large scales. Research efforts should focus on designing frameworks that can generalize from sparse supervisory signals, incorporating techniques like semi-supervised learning, transfer learning, and DG. These methods have the potential to unlock new applications in areas such as vegetation biomass estimation, urban morphology reconstruction, and hydrological modeling.  

\textbf{Datasets for LiDAR Remote Sensing.}
For large-scale environmental and geophysical parameter inversion, such as biomass estimation, LiDAR can serve as a critical weak supervisory signal \cite{weber2024unifieddeeplearningmodel,muszynski2024finetuninggeospatialfoundationmodels}. However, the lack of comprehensive, large-scale LiDAR datasets tailored to these tasks limits its potential. High-resolution LiDAR data integrated with other remote sensing modalities could bridge this gap, enabling more accurate and scalable models for remote sensing applications~ \cite{allen2024m3leo}.

In the domain generalization context, the absence of datasets facilitating cross-modal, cross-season and cross-platform transfer learning poses significant challenges. Current datasets are predominantly developed for autonomous driving and lack the scope and diversity required for Earth observation. For example, cross-modal datasets enabling the transfer of knowledge between space-borne, ALS, and TLS remain unavailable. Such datasets are essential for studying domain gaps and advancing DA techniques for varying acquisition geometries and resolutions—key challenges in remote sensing. Moreover, the scarcity of multimodal datasets combining LiDAR with optical, radar, or hyperspectral remote sensing data further constrains progress. Cross-modality datasets with aligned labels are critical for fostering WSL and enabling models to leverage complementary information across sensing modalities.

Future efforts should prioritize building large-scale, multimodal, and cross-platform LiDAR datasets for remote sensing. These datasets will form the foundation for developing weak supervision techniques and addressing core challenges in domain generalization, ultimately advancing the applicability of LiDAR in dynamic and large-scale Earth observation tasks.

\section{Conclusion} \label{section_conclusion}

\textcolor{black}{
In summary, this paper presents the first comprehensive overview of LiDAR remote sensing from a weakly supervised learning perspective. We revisit two core directions—data interpretation and parameter inversion—under a unified machine learning framework. For LiDAR data interpretation, we review key tasks such as semantic segmentation and object detection, emphasizing representative WSL techniques. For LiDAR-based inversion, we examine applications such as canopy and building height estimation, where weak supervision is emerging but remains underexplored. These interpretation and inversion tasks face significant domain shift across regions, seasons, and sensors, highlighting the importance of domain generalization within WSL frameworks for scalable LiDAR remote sensing.}

\textcolor{black}{ We further analyze LiDAR-specific challenges for WSL, such as irregular geometry and sampling sparsity, which demand tailored strategies beyond conventional 2D settings. Looking ahead, we outline several promising research directions: leveraging ubiquitous weak annotations, developing domain-adaptive WSL to mitigate domain shift, advancing inversion with new paradigms and datasets, and exploring foundation model integration to support open-world, scalable, and adaptive remote sensing. Through these explorations, we introduce an interdisciplinary perspective and conceptual framework that encourages the integration of machine learning, remote sensing, and geoscience for advancing LiDAR remote sensing.}

\section*{Acknowledgements}

This work was supported by the National Natural Science Foundation of China under Grant 42201481 and 42271365.

\bibliographystyle{elsarticle-num-names} 
\bibliography{ref}

\end{document}